\documentclass{article}

\usepackage{arxiv}

\usepackage[utf8]{inputenc}
\usepackage[T1]{fontenc}
\usepackage[scaled=.98]{XCharter}
\usepackage[scaled=.95]{helvet}
\usepackage[scaled=1.1]{zlmtt}
\usepackage{amsmath}
\usepackage[uprightscript,charter,vvarbb,scaled=1.05]{newtxmath}
\usepackage[hyphens]{url}
\usepackage{xcolor}
\definecolor{LinkBlue}{HTML}{2457C5}
\definecolor{AbstractGray}{HTML}{F3F5F7}
\definecolor{AbstractBorder}{HTML}{DDE3EA}
\usepackage[
  colorlinks=true,
  linkcolor=LinkBlue,
  citecolor=LinkBlue,
  urlcolor=LinkBlue
]{hyperref}
\usepackage{graphicx}
\usepackage{fontawesome5}
\usepackage{booktabs}
\usepackage{tabularx}
\usepackage{microtype}
\usepackage{enumitem}
\usepackage{etoolbox}
\usepackage[numbers,sort&compress]{natbib}
\usepackage{doi}
\usepackage[most]{tcolorbox}

\newtcolorbox{preprintabstract}{
  enhanced,
  breakable,
  colback=AbstractGray,
  colframe=AbstractBorder,
  boxrule=0.35pt,
  arc=3pt,
  left=12pt,
  right=12pt,
  top=5pt,
  bottom=9pt,
  before skip=10pt,
  after skip=16pt,
  title={Abstract},
  fonttitle=\large\bfseries\sffamily,
  coltitle=black,
  colbacktitle=AbstractGray,
  halign title=center,
  titlerule=0pt
}

\renewenvironment{abstract}
  {\begin{preprintabstract}\fontsize{10pt}{12.4pt}\selectfont}
  {\end{preprintabstract}}

\makeatletter
\renewenvironment{table}
  {\@float{table}}
  {\end@float}
\makeatother

\pdftrailerid{}

\newcommand{\appendixavailability}
  {Appendices~A--E are included in this document.}
\newcommand{\papercolumnfigurewidth}{0.51\textwidth}

\newcommand{\papertitle}{Fast Matrix Multiplication in fp8:\\
Certified Coefficient Optimization and Measured Error}
\newcommand{\papershorttitle}{Fast Matrix Multiplication in fp8}
\newcommand{\paperstatus}{Preprint}

\newcommand{\dezhicorrespondenceemail}{dezhiran@pku.edu.cn}
\newcommand{\taocorrespondenceemail}{taoxie@pku.edu.cn}
\newcommand{\artifacturl}{https://github.com/xieTwim/fast-matmul-fp8-artifact}
\newcommand{\artifactdisplay}{github.com/xieTwim/fast-matmul-fp8-artifact}

\title{\normalfont\sffamily\bfseries\papertitle}

\newcommand{\authornamefont}{\fontsize{11.8pt}{14.2pt}\selectfont\sffamily\bfseries}
\newcommand{\affiliationfont}{\fontsize{10.2pt}{12.4pt}\selectfont\normalfont}
\newlength{\authorrowbreakheight}
\makeatletter
\patchcmd{\@maketitle}
  {\begin{tabular}[t]{c}\bf\rule{\z@}{24\p@}\ignorespaces}
  {\begin{tabular}[t]{c}\authornamefont\rule{\z@}{24\p@}\ignorespaces}
  {}{\PackageWarning{arxiv-paper-template}{Could not patch the first author-row break}}
\patchcmd{\@maketitle}
  {\begin{tabular}[t]{c}\bf\rule{\z@}{24\p@}\ignorespaces}
  {\begin{tabular}[t]{c}\authornamefont\rule{\z@}{\authorrowbreakheight}\ignorespaces}
  {}{\PackageWarning{arxiv-paper-template}{Could not patch the second author-row break}}
\makeatother
\newcommand{\resourcelink}[2]{%
  \href{#1}{\normalfont\mdseries\nolinkurl{#2}}%
}
\newcommand{\resourceentry}[4]{%
  \raisebox{-0.05ex}{\makebox[1.35em][c]{\normalsize #1}}&
  \textbf{#2:}\ \resourcelink{#3}{#4}%
}

\author{
  \authornamefont
  \textbf{Shuxiao Xie}\textsuperscript{1,2}\thanks{Equal contribution.}
  \quad
  \textbf{Shuyang Xie}\textsuperscript{3}\footnotemark[1]
  \quad
  \textbf{Yuan Cao}\textsuperscript{1,4}
  \AND
  \textbf{Dezhi Ran}\textsuperscript{1}\thanks{Corresponding authors:
  \resourcelink{mailto:\dezhicorrespondenceemail}{\dezhicorrespondenceemail} and
  \resourcelink{mailto:\taocorrespondenceemail}{\taocorrespondenceemail}.}
  \quad
  \textbf{Wei Yang}\textsuperscript{2}
  \quad
  \textbf{Tao Xie}\textsuperscript{1,2,4,5}\footnotemark[2]
  \\[0.7em]
  \affiliationfont
  \textsuperscript{1}Beijing Tongming Lake Information Technology Application
  Innovation Center (TLAIC), China
  \\
  \textsuperscript{2}Fudan University Institute of Systems for Advanced Computing, China
  \\
  \textsuperscript{3}Harbin Institute of Technology, China
  \\
  \textsuperscript{4}Key Lab of HCST (PKU), MOE; SCS, Peking University, Beijing, China
  \\
  \textsuperscript{5}Shanghai Institute of Systems for Open Computing, China
}

\date{\small
\begin{tabular}{@{}r@{\hspace{0.4em}}l@{}}
\resourceentry{\faGithub}{Reproducibility artifact}{\artifacturl}{\artifactdisplay}
\end{tabular}
}

\renewcommand{\headeright}{\paperstatus}
\renewcommand{\undertitle}{\paperstatus}
\renewcommand{\shorttitle}{\papershorttitle}

\hypersetup{
  pdftitle={Fast Matrix Multiplication in fp8: Certified Coefficient Optimization and Measured Error},
  pdfsubject={Public preprint},
  pdfauthor={Shuxiao Xie, Shuyang Xie, Yuan Cao, Dezhi Ran, Wei Yang, Tao Xie},
  pdfkeywords={fast matrix multiplication, fp8, global optimality certificate,
    tensor decomposition, low-precision arithmetic}
}

\begin{document}
\maketitle

\begin{abstract}
A Strassen-type algorithm has many realizations with the same exact product and multiplication count yet different fp8 error because basis changes reshape coefficient geometry, posing the question of which to run. No current account settles this: classical stability controls worst-case \(\ell_1\) growth, not the expected-error magnitude, and the Dumas--Pernet--Sedoglavic optimizer could only be called probably optimal, its global optimality unproved. To settle this, we attach to each realization a coefficient functional \(\Phi\), a scalar summary of its coefficient geometry, which we minimize over the change-of-basis orbit. This Kempf--Ness problem on a Hadamard manifold lets us certify the global \(\Phi\) optimum rather than merely search for it: an exact moment-map zero fixes \(\Phi_{\min} = 200/9\), and de Groote's classification extends that optimality to every exact real rank-7 \(2\times2\) decomposition. Every exact real rank-7 realization therefore has a \(\Phi\)-predicted RMS constant at least \(5/3\) times that of the cubic algorithm, at fixed noise coefficient. We then introduce an explicit block-scaled e4m3 model in which \(\Phi\) is the leading-order coefficient of relative expected mean-squared error, and we test the resulting \(\Phi\)-predicted ordering against realized fp8 error. Ordering and re-basing experiments support that prediction within tested fused block-scaled regimes, and on real matmul tiles from four architecture families the \(\Phi\)-optimal realization falls in the fp8 low-error region. Across two \(\sim\)70B models on real \texttt{deep\_gemm} kernels, the same realization removes 10 to 55\% of classic Strassen's excess NLL over the clean model. Algorithm realization thus becomes a mathematically certified design problem rather than a tuning choice: an independent low-precision axis with a global \(\Phi\) optimum and measured fp8 relevance.
\end{abstract}

\providecommand{\appendixavailability}
  {Appendices~A--E form a separate technical appendix, submitted as supplementary material.}
\providecommand{\papercolumnfigurewidth}{\columnwidth}

\section{Introduction and Positioning}

As large-model inference and training have scaled, there is growing interest in whether Strassen-type fast matrix multiplication can be put to practical use \citep{falcongemm, subcuber}, while fp8 has become common for these workloads. A fast matrix-multiplication algorithm can be written in many equivalent coordinate forms related by a change of basis, all with the same exact product and multiplication count.

\begin{figure}[t]
\centering
\includegraphics[width=\papercolumnfigurewidth]{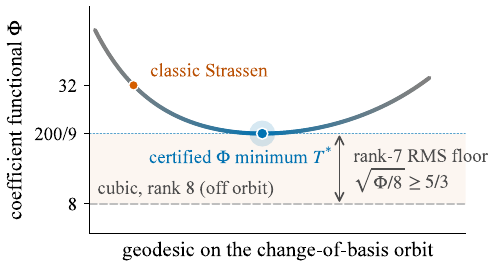}
\caption{A geodesic on the change-of-basis orbit of a fixed rank-7 \(\langle 2,2,2\rangle\) decomposition, with the coefficient functional \(\Phi\) as height. Equivalent realizations share the tensor and multiplication count but differ in \(\Phi\), hence in RMS scale under the block-scaled fp8 error model. Cubic multiplication has rank 8 and is a reference level, not a point on this orbit.}
\label{fig:orbit-teaser}
\end{figure}

Those forms are distinct algorithm realizations, and they are not equally accurate. A change of basis rewrites the \mbox{coefficient} vectors each rank-one product uses to encode its two operands and decode its result. Block-scaled fp8 quantizes the encoded combinations, not the original operands, so changing a term's encoding vectors changes the rounding noise it injects. The two encoder norms set the injected variance and the decoder norm its weight in the output; a term's leading expected-error contribution therefore scales as the product of its three squared coefficient norms. Summing over terms gives a single scalar, the coefficient functional \(\Phi\), that a change of basis moves while the exact product and multiplication count stay fixed (Figure~\ref{fig:orbit-teaser}). A basis is thus not a coordinate convention but an independent low-precision design axis. This paper settles which realization minimizes \(\Phi\), and tests how far the resulting ordering carries into realized fp8 error.

That question sits closest to numerical-stability theory, but existing analyses stop short of the algorithm-realization objective. Classical work bounds worst-case \(\ell_1\) coefficient growth or related normwise factors rather than the expected error induced by block-scaled fp8 rounding \citep{higham-1990, demmel-ddhk-2007, ballard-2016, dai-lim-2023}. Closest to our setting, constructions using alternative bases or stability optimization vary the same coefficient degrees of freedom but target arithmetic cost and worst-case error bounds, not block-scaled expected squared error \citep{schwartz-alt-basis, vermeylen-2023}. Dumas, Pernet, and Sedoglavic (DPS) minimize a related \(\ell_1\)-type factor over the same change-of-basis orbit, but describe their minimizer only as ``probably optimal,'' leaving a gap of at most 2.6\% open \citep{dumas-strassen-2024, dumas-pernet-sedoglavic-2025}. Mary's Research Problem~16.2 identifies the corresponding gap on the error side: a probabilistic error analysis for Strassen-type algorithms, for which standard probabilistic rounding-error tools do not directly apply. It points to accuracy-oriented variants of Dumas et al.\ as a target \citep{mary-2025}. Narrow-range error analyses treat the format's underflow and scaling behavior rather than the choice among equivalent realizations \citep{mary-narrow-2025}, and flip-graph searches discover new schemes over discrete coefficient rings \citep{perminov-2026}, a different axis from the continuous change of basis within one fixed exact class. Operand-side methods such as SmoothQuant, QuaRot, and SpinQuant address a complementary axis because they change the operands rather than the multiplication algorithm \citep{smoothquant, quarot, spinquant, ordentlich-2026, ordentlich-highrate-2026}. For the realization choice posed above, both the block-scaled expected-error functional and its global optimum therefore remain unsettled.

We settle both by turning the realization choice into geometry. Minimizing \(\Phi\) over a change-of-basis orbit is a Kempf--Ness problem \citep{bfgoww-2019}, and that structure is what lets us certify a minimum instead of merely searching for one. The orbit is unbounded, so no finite sweep can exclude a better basis, but once the compact symmetries are quotiented away the remaining variables live on a Hadamard manifold along whose geodesics \(\Phi\) is convex. Convexity turns a first-order condition into a global one. That condition is met symbolically at the minimizing representative, the one whose three coefficient vectors carry equal norms in every term and at which the moment map vanishes exactly. It fixes \(\Phi_{\min}=200/9\) for Strassen's rank-7 \(2\times2\) orbit \citep{strassen1969}. By de Groote's classification every exact real rank-7 algorithm for \(2\times2\) multiplication lies in that orbit, so the certificate holds class-wide as a corollary \citep{degroote-1978}. The same representative is also the exact global minimizer of the classical \(\ell_1\) factor, which closes the gap DPS left open.

Certifying the optimum of \(\Phi\) is worth doing only if fp8 error follows \(\Phi\). We therefore make that link explicit rather than assumed, deriving a second-moment model for fused block-scaled fp8, the regime in which operands are encoded, quantized per block, and accumulated in a single pass. Under that model \(\Phi\) is the leading-order coefficient of expected mean-squared error. The model gives the certificate its fp8 meaning and is where the argument becomes an approximation rather than a theorem. Section~4 tests the resulting \(\Phi\)-predicted ordering rather than assuming it, and the tests come out in its favor. Re-basing a decomposition to its \(\Phi\)-minimizing form lowers measured fp8 error in every one of the tested decompositions, by at least \(1.14\times\). At model scale, the \(\Phi\)-optimal realization lowers accumulated loss relative to classic Strassen on real \texttt{deep\_gemm} kernels. These outcomes support the prediction within the tested fused block-scaled regimes.

Our work makes three contributions.
\begin{itemize}
\item \textbf{A global optimality certificate for fast-matmul realization.} We establish a global \(\Phi\)-optimality certificate for the \(\langle 2,2,2;7\rangle\) change-of-basis orbit and, through de Groote's classification, extend it to the entire exact real rank-7 class; the same argument closes DPS's \(\ell_1\) gap. The certificate is a class-wide lower bound, hence a no-free-lunch floor: at fixed noise coefficient, no exact real rank-7 realization has a \(\Phi\)-predicted RMS error constant below \(5/3\) times cubic multiplication's.
\item \textbf{A scoped second-moment bridge from the certificate to fp8 error.} We derive the second-moment model that makes \(\Phi\) the leading-order coefficient of expected fp8 error, converting an exact statement about coefficient geometry into a \(\Phi\)-predicted ordering of realizations.
\item \textbf{Evidence that the \(\Phi\)-optimal realization matters in real kernels.} We show that on real matmul tiles from five models across four architecture families the \(\Phi\)-optimal realization falls in the low-error region of the real-kernel landscape, beating classic Strassen on \(98.4\%\) of those tiles, averaged equally over the models. On two roughly 70B-parameter models running real \texttt{deep\_gemm} kernels it removes 10.2\ to 55.0\% of classic Strassen's excess next-token loss over the clean model.
\end{itemize}
Fast-matmul realization thus becomes a certified coefficient-geometry problem whose fp8 relevance is measured and delimited rather than assumed. \appendixavailability

\section{The Orbit Objective and Its fp8 Interpretation}

\subsection{Orbit Geometry and the Coefficient Functional}

The realization choice from Section~1 is formalized by a fixed tensor decomposition and its change-of-basis orbit. A bilinear algorithm \(\langle m,k,n;R\rangle\) is a rank-\(R\) CP decomposition of the matrix-multiplication tensor. Fixed vectorizations \(a,b,c\) of the input and output blocks give
\[
\begin{gathered}
\mathcal{M}_{m,k,n}=\sum_{r=1}^{R}U_r\otimes V_r\otimes W_r,\\
c=\sum_{r=1}^{R}\langle U_r,a\rangle\langle V_r,b\rangle W_r,
\end{gathered}
\]
where term \(r\) carries coefficient vectors \((U_r,V_r,W_r)\). An invertible triple \((X,Y,Z)\in\mathrm{GL}(m)\times\mathrm{GL}(k)\times\mathrm{GL}(n)\) acts by a compatible basis change on operand matrices \(A,B\) and output \(C\) (vectorized above as \(a,b,c\)), sending them to \(XAY^{-1}\), \(YBZ^{-1}\), and \(XCZ^{-1}\) and re-expressing each coefficient triple \((U_r,V_r,W_r)\) accordingly. For a decomposition \(D\) and \(g=(X,Y,Z)\), denote the result by \(D\cdot g\); this is a right action, so \(D\cdot(gh)=(D\cdot g)\cdot h\). This action preserves the exact product and multiplication count \(R\). These decompositions represent the change-of-basis orbit.

For \(D=\{(U_r,V_r,W_r)\}_{r=1}^{R}\), the fp8 model selects the coefficient functional:
\[
\Phi(D)=\sum_{r=1}^{R}\lVert U_r\rVert_2^2\,\lVert V_r\rVert_2^2\,\lVert W_r\rVert_2^2.
\]
Within this orbit, we treat three \(\Phi\)-invariant changes as gauge: an orthogonal sandwich, term permutation, and product-one termwise rescaling \((U_r,V_r,W_r)\mapsto(\alpha_rU_r,\beta_rV_r,\gamma_rW_r)\), \(\alpha_r\beta_r\gamma_r=1\); the latter two only alter the representation. For \(\langle2,2,2\rangle\), the orthogonal factor is \(\mathrm{O}(2)^3\); its finite stabilizer within the factor-preserving sandwich action is isomorphic to \(S_3\), acts through the listed term-permutation and rescaling identifications, and adds no separate gauge. Quotienting this orbit by these gauges gives the gauge quotient below, on which \(\Phi\) is well defined. Let \(\lambda_r=\lVert U_r\rVert_2\lVert V_r\rVert_2\lVert W_r\rVert_2\) denote the gauge-invariant rank-one magnitude of term \(r\). Then
\[
\begin{gathered}
\Phi_\kappa(D)=\sum_{r=1}^{R}\lambda_r^{\kappa},\\
\Phi=\Phi_2=\lVert\lambda\rVert_2^2,\qquad
\Gamma=\Phi_1=\lVert\lambda\rVert_1 .
\end{gathered}
\]
The fixed value \(\kappa=2\) gives this second-moment, or \(\ell_2\), form; classical stability theory instead controls the worst-case \(\ell_1\) growth factor \(\Gamma\), so these identities identify \(\Phi\) as its probabilistic \(\ell_2\) counterpart. Thus minimizing \(\Phi\) compares realizations at fixed exact product and \(R\). A general non-orthogonal sandwich is genuine orbit freedom rather than gauge: it can change \(\Phi\) and supplies the continuous freedom we optimize.

\subsection{The fp8 Error Interpretation}

Its fp8 meaning follows from a second-moment model. \emph{Assumption 1 (iid decorrelated operand-noise model).} Both operands have iid zero-mean entries. The original operand blocks remain unquantized. For each term \(r\), the encoded combinations \(x_r=\langle U_r,a\rangle\) and \(y_r=\langle V_r,b\rangle\) are formed first and then quantized. Their block-scaled e4m3 \citep{fp8-formats} operand-rounding noises are zero mean, with per-block operand-rounding noise coefficient \(s_q\), a normalized RMS of the operand rounding relative to the encoded-block scale. These noises are operand-independent and mutually uncorrelated across the two operand sides, terms \(r\), and inner coordinates. Let \(\epsilon_{x,r}\) and \(\epsilon_{y,r}\) denote the operand-rounding errors. Linearizing \((x_r+\epsilon_{x,r})(y_r+\epsilon_{y,r})\) leaves \(\epsilon_{x,r}y_r+x_r\epsilon_{y,r}\). The decoder \(W_r\) carries this error into the vectorized output as \((\epsilon_{x,r}y_r+x_r\epsilon_{y,r})W_r\), contributing the factor \(\lVert W_r\rVert_2^2\) to its second moment. Under Assumption~1, the covariance between the two operand-side errors and all covariances with \(r\neq r'\) vanish. Each encoded operand's rounding variance is \(s_q^2\) times its energy, so \(\epsilon_{x,r}y_r\) has second moment proportional to \(s_q^2\lVert U_r\rVert_2^2\lVert V_r\rVert_2^2\), as does \(x_r\epsilon_{y,r}\); together they sum to twice one side. Term \(r\) therefore contributes a quantity proportional to \(2s_q^2\lVert U_r\rVert_2^2\lVert V_r\rVert_2^2\lVert W_r\rVert_2^2\). Summing over \(r\) gives \(2s_q^2\Phi(D)\) times the product of the two operand second moments. With \(mn\) output blocks each summing \(k\) inner products, the exact-product second moment \(\mathbb{E}\lVert c\rVert_2^2\) equals \(mkn\) times that same operand-moment product, up to the per-block entry factor \(PQL\) from Appendix~D; both cancel under normalization. Writing \(\delta c\) for the output error gives
\[
\frac{\mathbb{E}\lVert \delta c\rVert_2^2}{\mathbb{E}\lVert c\rVert_2^2}\;\approx\;\frac{2\,s_q^2\,\Phi(D)}{mkn}.
\]
Together, Assumption~1 and this leading-order relation define the second-moment model, in which \(\Phi\) is the leading-order coefficient of the relative expected mean-squared error. Appendix~D gives the full derivation.

Assumption~1 makes \(\Phi\) the fp8-relevant target at leading order; Section~3 certifies its exact class-wide minimum over exact real rank-7 \(\langle 2,2,2\rangle\) realizations, and Section~4 tests the resulting \(\Phi\)-predicted ordering in measured fp8 error within operand families.

\section{A Class-Wide Global Optimality Certificate}

\subsection{Orbit Minimality and Its Class-Wide Extension}

\noindent\textbf{Theorem 1 (class-wide $\Phi$ optimum).} For every exact real rank-$7$ decomposition $D$ of $\langle 2,2,2\rangle$, at fixed $\kappa=2$,
\[
\min_{g\in\mathrm{GL}_2(\mathbb R)^3}\Phi(D\mathbin{\cdot}g)=\Phi_{\min}=200/9,
\]
and, modulo the $\Phi$-preserving gauge, the minimizing class is represented by the balanced rank-$7$ decomposition $T^*$. Here $T^*$ is Strassen's decomposition under the balanced sandwich $X=Y=Z=T_\rho$, where, in our right-action convention, $T_\rho=\left(\begin{smallmatrix}\rho^{-1}&\rho/2\\0&\rho\end{smallmatrix}\right)$ is the inverse of the minimizing DPS Proposition~28 factor at $\rho=(4/3)^{1/4}$ \citep{dumas-pernet-sedoglavic-2025}. The proof requires two steps: first, a Kempf--Ness reduction minimizes $\Phi$ over this representative's change-of-basis orbit after quotienting the compact symmetries; second, de Groote's classification identifies that orbit with the entire exact real rank-$7$ class.

The first step proves the orbit minimum for $T^*$. A $\Phi$-neutral determinant-one normalization first removes the central scalings, whose contributions cancel exactly. For each normalized transformation, polar decomposition then gives $g=P\mathbin{\cdot}Q$, with $P$ symmetric positive definite and $Q\in\mathrm{O}(2)^3$. The order matters under the right action:
\[
T^*\mathbin{\cdot}g=(T^*\mathbin{\cdot}P)\mathbin{\cdot}Q.
\]
Thus the last-acting orthogonal factor leaves $\Phi$ unchanged, while the noncompact variables descend to $(\mathrm{SL}_2(\mathbb R)/\mathrm{SO}_2)^3$, a complete, simply connected, nonpositively curved Hadamard manifold. There, $\Phi$ is the squared orbit norm. Along every symmetric-space geodesic, the decomposition splits into weight components with exponentially varying squared magnitudes, making $\Phi$ a finite sum of exponentials with nonnegative coefficients and thus convex along the entire path, not merely near $T^*$. Because $T^*$ is polystable, its orbit is closed, so this globally convex problem attains its infimum. The displayed $T_\rho$ fixes the common sandwich matrix exactly, hence the coordinates of $T^*$, and at those coordinates the moment map satisfies the symbolic identity $\mu(T^*)=0$. Geodesic convexity and this exact first-order balance prove that $T^*$ is a global $\Phi$ minimizer on its orbit. A symbolic script checks the zero exactly and rejects a deliberately perturbed negative control. Table~\ref{tab:phi-certificate} records each component's logical role and evidence, and Appendix~A gives the coordinate expansion and full moment-map arithmetic.
\begin{table*}[t]
\centering
{\small
\begin{tabularx}{\textwidth}{@{}>{\raggedright\arraybackslash}p{0.46\textwidth}>{\raggedright\arraybackslash}X@{}}
\toprule
Component --- logical role & Evidence \\
\midrule
$\Phi(T^*)=200/9$ \newline \hspace*{1em}{\itshape Value at the candidate} & Exact (symbolic) \\
\addlinespace[3pt]
$\mu(T^*)=0$ \newline \hspace*{1em}{\itshape Stationarity; globality when combined with convexity} & Exact (symbolic): $\mathcal L_A=\mathcal R_A=\tfrac{100}{9}I_2$ for $A=U^*,V^*,W^*$; perturbed negative control rejected \\
\addlinespace[3pt]
$\Phi$ geodesically convex on the quotient \newline \hspace*{1em}{\itshape Globality over the orbit} & Analytic proof (Appendix~A) \\
\addlinespace[3pt]
Second-order form positive definite off gauge \newline \hspace*{1em}{\itshape Strictness and uniqueness on the gauge quotient} & Exact (symbolic; eigenvalues $128/9$ and $440/9$, multiplicities two and four) \\
\addlinespace[3pt]
de Groote's classification \newline \hspace*{1em}{\itshape Lifts the orbit minimum to the entire exact real rank-$7$ class} & Cited theorem; reductions in Appendix~B \\
\bottomrule
\end{tabularx}}
\caption{Components of the $\Phi$-optimality certificate at $T^*$. The first four are orbit-level; the last carries the orbit result to the whole class.}
\label{tab:phi-certificate}
\end{table*}

Within this orbit, after adding only the $\Phi$-neutral gauges, these facts determine the equality set. An equivalent decomposition has $\Phi=200/9$ if and only if it is obtained from $T^*$ by an orthogonal sandwich in $\mathrm{O}(2)^3$, a term permutation, and an admissible product-one termwise rescaling. Hence $T^*$ is the unique minimizer on this orbit's gauge quotient. The symbolic restricted second-order matrix is positive definite on the span of all six noncompact directions, the full tangent complement of the gauge directions. It therefore establishes strictness and uniqueness there, leaving no zero mode or unchecked subspace; globality follows from geodesic convexity and $\mu(T^*)=0$. If a second minimizing class existed, choose representatives $y$ of that class and $x^*$ of $T^*$ in the Hadamard manifold $(\mathrm{SL}_2(\mathbb R)/\mathrm{SO}_2)^3$. Because the classes are distinct, $y$ and $x^*$ are not related by the finite stabilizer of the base-point decomposition, so the unique geodesic joining them has a nonzero nongauge initial tangent at $x^*$. Since both endpoints are global minima and $\Phi$ is convex along this geodesic, $\Phi$ is constant on the segment from $x^*$ to $y$; the second derivative of $\Phi$ along it at $x^*$ therefore vanishes. This contradicts the positive-definite restricted Hessian along that nonzero nongauge direction, so uniqueness descends to the gauge quotient. The finite base-point stabilizer adds no equality case because it acts through the listed term-permutation and rescaling gauges. Appendix~A gives the full restricted matrix and its positive-definiteness evidence.

The second step lifts this orbit statement to the whole exact real rank-$7$ class. De Groote's 1978 Theorem~0.1(i) states that every exact real rank-$7$ decomposition of $\langle 2,2,2\rangle$ is equivalent to Strassen's through a real sandwich $g\in\mathrm{GL}_2(\mathbb R)^3$, a term permutation in $\mathfrak S_7$, and termwise rescaling \citep{degroote-1978}; see also \citep{burichenko-2015}. Since $T^*$ lies in Strassen's orbit, de Groote's theorem implies that the first-step orbit, after adding the $\Phi$-neutral permutation and admissible-rescaling gauges, covers the entire exact real rank-$7$ class. It follows that every such decomposition satisfies $\Phi(D)\geq 200/9$, with equality precisely for the orthogonal, permutation, and admissible-rescaling gauge class of $T^*$. The minimization retains the real-sandwich change-of-basis action, which may change $\Phi$ and is not an invariance. The remaining CP indeterminacies are $\Phi$-neutral analytic identities. Writing
\[
t_r=U_r\otimes V_r\otimes W_r,
\qquad
\Phi=\sum_r\lVert t_r\rVert_{\mathrm{HS}}^2,
\]
expresses the functional as the sum of squared Hilbert--Schmidt norms of the rank-one tensors, independently of how each $t_r$ is factorized. Admissible product-one factor rescalings therefore leave each $t_r$ unchanged, term permutations merely reorder the sum, trailing orthogonal sandwiches preserve the Hilbert--Schmidt norms, and central-scalar contributions cancel exactly. The class-wide result is therefore a corollary of de Groote combined with the exact base-point algebra and the analytic orbit argument. Appendix~B gives the invariance reductions and the real-field step.

\subsection{Quantitative Consequences and the \texorpdfstring{$\Gamma$}{Gamma} Optimum}

The cubic algorithm has $\Phi=8$, distinct in meaning from the coincident dimension product $mkn=8$. Therefore
\[
\Phi_{\min}/\Phi_{\mathrm{cubic}}=(200/9)/8=25/9,
\]
and the square root $5/3$ is the $\Phi$-predicted relative-RMS coefficient. At fixed noise coefficient, every exact real rank-$7$ realization consequently has a $\Phi$-predicted RMS coefficient at least $5/3$ times cubic's. This is the functional's $5/3$ no-free-lunch floor. Classic Strassen has $\Phi=32$, so its $\Phi$-predicted RMS coefficient is exactly twice cubic's; re-basing to $T^*$ moves a rank-7 realization from twice cubic's predicted RMS constant down to the floor. Under Assumption~1 and at fixed noise coefficient, this exact $\Phi$ floor induces the corresponding fp8-model floor on the leading-order relative expected mean-squared-error coefficient, whose square root is the $5/3$ relative-RMS coefficient.

The coefficient functional is exactly multiplicative under Kronecker products, $\Phi(A\otimes B)=\Phi(A)\Phi(B)$; consequently the Kronecker square of $T^*$ is a $\langle 4,4,4;49\rangle$ realization with $\Phi=(200/9)^2=40000/81$, and at Kronecker-power, or recursion, depth $\ell$ the $\Phi$-predicted RMS coefficient propagates as $(5/3)^\ell$. This is a $\Phi$-optimality certificate for the composite orbit---the image under the Kronecker embedding of the product of $\ell$ base-level sandwich groups, consisting of transformations of the form $g_1\otimes\cdots\otimes g_\ell$ rather than the full $\mathrm{GL}_{2^\ell}(\mathbb R)^3$---not a global $\langle 4,4,4\rangle$ optimum.

The classical $\ell_1$ growth factor $\Gamma$ has geodesic convexity distinct from that established above for $\Phi$. The DPS Eq.~(98) representative $T^*$ is an exact $\Gamma$ critical point. Global geodesic convexity together with this exact criticality implies that $T^*$ is the exact global minimizer, with $\Gamma_{\min}=2\sqrt2+16/\sqrt3$, and closes the $\leq 2.6\%$ gap left open by DPS \citep{dumas-pernet-sedoglavic-2025}. On $\langle 2,2,2;7\rangle$, the $\Gamma$- and $\Phi$-optima thus coincide. Relative to $\Gamma$, $\Phi$ supplies the second-moment magnitude, the $\ell_2$ rather than $\ell_1$ geometry for ordering realizations, and the $5/3$ floor. Appendix~C proves $\Gamma$-globality analytically from geodesic convexity and exact criticality; the executable check verifies the exact value. The standing result is class-wide: across the entire exact real rank-$7$ $\langle 2,2,2\rangle$ class, $T^*$ is the unique $\Phi$-optimal realization \mbox{modulo} gauge. Section~4 tests whether this $\Phi$-predicted ordering appears in measured fp8 error.

\providecommand{\papercolumnfigurewidth}{\columnwidth}

\section{Empirical Tests of the Second-Moment Model}

Section~3 fixes the coefficient optimum exactly; whether that optimum is the realization to run is a physical question. The second-moment model makes one prediction: realizations rank by \(\Phi\) at fixed shape, and by \(\Phi/mkn\) across shapes. This section tests that prediction progressively farther from its derivation: to other orbits, to other shapes, then to real operands and a real kernel, and finally to accumulated model loss.

Two conventions hold throughout. \emph{Fused} names Section~1's single-pass regime, the one used to derive the model. The exact certificate of Section~3 covers the full \(\langle 2,2,2;7\rangle\) orbit and its class, together with the composite orbits of its Kronecker powers; general-shape optima below are therefore numerical and named \(\Phi\)-minimizing rather than certified.

\subsection{Emulated Ordering and Re-basing}

The first question is whether \(\Phi\) orders realizations. With the shape fixed at \(\langle 2,2,2;7\rangle\), so that \(mkn\) is constant and \(\Phi\) alone varies, \(\Phi\) orders mean end-to-end KL across eighteen realizations on Qwen2.5-7B with Spearman correlation \(0.9850\). That result could still be specific to the one orbit the certificate was proved on, so the next test leaves it: over six re-based \(\langle 4,4,4\rangle\) points, one rank-48 and five rank-49, the tie-aware correlation between \(\Phi/mkn\) and emulated error is \(0.986\).

Neither probes the model's derived form: with \(mkn\) held fixed, the \(1/mkn\) normalization has no effect, and only varying the shape puts it at risk. Across twenty AlphaTensor shapes \citep{alphatensor} the model-consistent predictor \(\sqrt{\Phi/(mkn)}\) reaches \(0.9940\) against measured relative Frobenius error, while bare \(\sqrt{\Phi}\), without that normalization, reaches \(0.9455\). The coefficient-growth heuristic we adapt from DPS reaches \(0.9759\) on the same shapes \citep{dumas-pernet-sedoglavic-2025}. The bare comparison is informative: the ranking supports the model's derived form, not a general preference for smaller coefficients; Appendix~E defines the heuristic and gives the per-shape table.

\begin{figure*}[t]
\centering
\includegraphics[width=\textwidth]{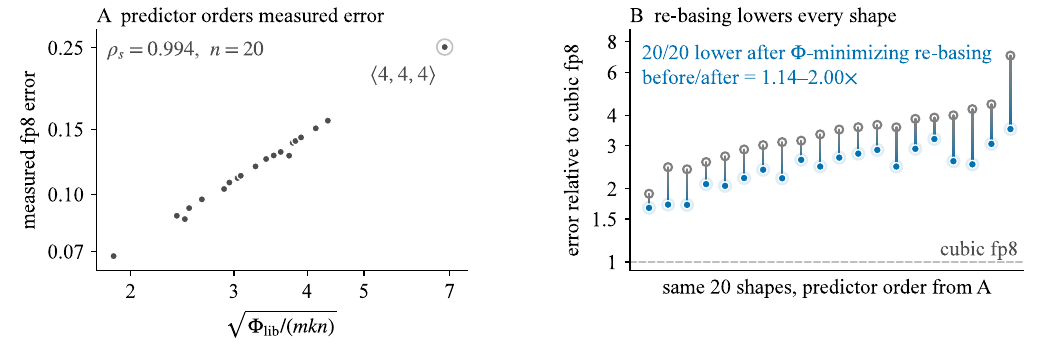}
\caption{Prediction and intervention over twenty AlphaTensor shapes, using an fp32 emulator of fused block-scaled fp8 with blk128 blocks on one signed-uniform synthetic operand family. Panel A plots one library representative per shape: the coefficient-only predictor \(\sqrt{\Phi_{\mathrm{lib}}/(mkn)}\), which uses no measured error, against measured relative Frobenius error on logarithmic axes; \(\langle 4,4,4\rangle\) marks a poor library representative. Panel B pairs the same decompositions before re-basing (hollow) and after (filled), relative to a cubic-fp8 reference, on a logarithmic vertical axis. These general-shape minima are numerical, not certified.}
\label{fig:predict-intervene}
\end{figure*}

\begin{figure}[t]
\centering
\includegraphics[width=\papercolumnfigurewidth]{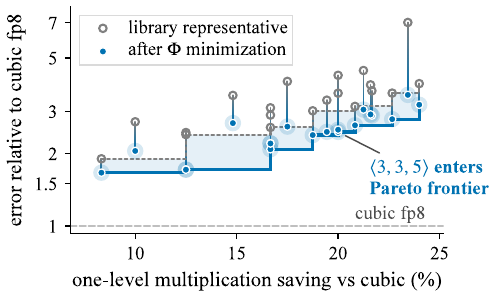}
\caption{Multiplication-count/error frontier for the same twenty shapes as Figure~\ref{fig:predict-intervene}. This re-plots those interventions under the same emulated setup; it is not a second experiment. Each vertical pair holds shape and multiplication count fixed. Step curves show the Pareto frontiers before and after numerical \(\Phi\) minimization on a logarithmic vertical axis.}
\label{fig:frontier}
\end{figure}

Correlation alone does not show improvement from optimizing the predictor, so we intervene. Re-basing each library realization to its \(\Phi\)-minimizing form lowers fp32-emulated fused-blk128 error in all \(20\) tested decompositions, by at least \(1.14\times\), on one signed-uniform synthetic operand family with one re-basing orbit per decomposition. Figure~\ref{fig:predict-intervene} juxtaposes the cross-shape prediction and paired intervention. Because re-basing holds the shape and multiplication count fixed, it adds no multiplications, and Figure~\ref{fig:frontier} shows the same twenty pairs as a multiplication-count/error frontier: the interventions move the nondominated set, not merely lowering twenty measurements, with \(\langle 3,3,5\rangle\) entering the frontier after re-basing. One comparison survives real operands: over 196 Qwen2.5-7B layer-by-type cells, on real activations and weights with emulated arithmetic, the \(\Phi\)-optimal realization has lower error than classic Strassen in 195.

\begin{figure}[t]
\centering
\includegraphics[width=\papercolumnfigurewidth]{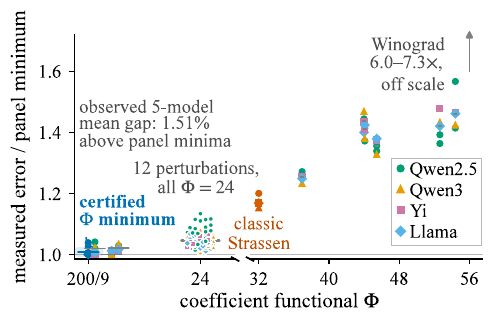}
\caption{Real-operand panel-gap test over the fixed twenty-four-arm orbit panel. Each marker is one model's energy-weighted per-matmul error across its share of the \(2352\) real \texttt{deep\_gemm} observations, divided by that model's own panel minimum. Marker shape encodes architecture family; the five models, including two from one family, are weighted equally. Short horizontal bars are across-model medians, except at \(\Phi=24\), where the bar pools all twelve arms; the ladder's fine ordering is not resolved. The horizontal axis is expanded on the left and broken across the arm-free interval; the twelve arms at \(\Phi=24\) are packed around that value, and arms sharing a lower \(\Phi\) are offset slightly. Winograd is off scale.}
\label{fig:oracle-gap}
\end{figure}

\subsection{Real Operands and the Real Kernel}

Everything above is emulated. Real deployment departs from the derivation in three ways at once: activations are not iid, the kernel scales weights in \(128\times128\) blocks where the emulator uses \(1\times128\) on both sides, and its single pass need not honor the noise independence the derivation assumes. The next test gives up all three. We measure \(2352\) real-operand observations, or tiles, each an activation-by-weight matmul for one linear layer and one context. They span five models across four architecture families: Qwen2.5-7B and Qwen2.5-32B from one family together with Qwen3-8B, Yi-6B, and Llama-3.3-70B-Instruct \citep{qwen25,qwen3,yi,llama3}, all run on the real \texttt{deep\_gemm} blk128 kernel. The comparison runs over a fixed panel of twenty-four \(\langle 2,2,2;7\rangle\) realizations, or arms, in the canonical rebalanced gauge, a \(\Phi\)-neutral normalization that equalizes the two encoder norms in every term; realized fp8 error, unlike \(\Phi\), is not invariant to that split. Twenty arms were pre-registered; four third-factor directions were added before the first sweep, after which the panel was frozen. Eighteen arms form a ladder ordered by \(\Phi\), containing six named realizations and twelve local \(\pm\delta\) perturbations of the optimum; the remaining six form three adversarial \(\Gamma\)-tie/\(\Phi\)-split pairs. Evidence from this panel is local to its twenty-four arms, not the continuous orbit, and per-matmul, not end-to-end.

Direction survives the transfer; resolution does not. Across cubic and three named panel arms the \(\sqrt{\Phi}\)-predicted order is cubic \(<\) \(\Phi\)-optimal \(<\) classic Strassen \(<\) Winograd \citep{dai-lim-2023}, and \(0.987\) of tiles show no inversion among them, after excluding pairwise comparisons whose measured errors tie within three percent and averaging equally over the five models. Within the eighteen-arm ladder, where the realizations sit close together in \(\Phi\), the per-tile Spearman correlation falls to \(0.59\)--\(0.64\) across the five per-model summaries, and the pre-registered fine-ordering O4 gate fails. Figure~\ref{fig:oracle-gap} shows both scales: the coarse rise of error with \(\Phi\) is resolved, while the ladder's internal order is not. The real kernel therefore reproduces the model's ordering of distinct realizations far more reliably than its fine spacing, which is the resolution the next test has to be read against.

Ordering is one thing; whether the \(\Phi\)-optimal realization sits where the error is lowest is another. Weighting each tile by its output energy within a model and then averaging equally across models, it lands within a relative panel gap \(E(\mathrm{opt})/E(\min)-1\) of \(0.0151\) of the best arm in the panel, and beats all twelve signed \(\pm\delta\) perturbations at once on a five-model mean fraction \(0.741\) of tiles. It is the exact panel minimum on Llama-3.3-70B and within \(0.27\%\) of it on Yi-6B; on the two Qwen2.5 models and on Qwen3-8B the panel minimum is instead a nearby DPS variant at \(\Phi\times1.019\), which beats it by \(2.63\%\) and \(3.85\%\) on the two Qwen2.5 sizes. That separation sits inside the ladder, where the previous test already found resolution limited. Resampling layer clusters bears that out: weighting the models equally, the optimum trails that variant by \(1.06\%\), with a 95\% interval of \([-0.76,3.45]\%\) that five model clusters leave coarse. The panel therefore does not separate the two arms, while the per-model intervals do, on the two Qwen2.5 sizes in the variant's favor and on Llama-3.3-70B in the optimum's.

The adversarial pairs ask the sharper question because the classical theory and ours give opposite rankings by construction: each nearly ties the \(\ell_1\) growth factor \(\Gamma\) while separating \(\Phi\), and the small residual \(\Gamma\) difference points the opposite way. Real error follows \(\Phi\) on a five-model mean \(0.862\) of separated tiles across the three pairs, so where the two functionals order a pair differently, the kernel follows the second-moment one rather than the worst-case one \citep{dumas-pernet-sedoglavic-2025} in the large majority of cases.

Against classic Strassen at identical rank the optimum has lower per-matmul error on \(98.4\%\) of tiles, again averaged equally over the five models, with a mean of per-model median reductions of \(14.8\%\). The five per-model medians of the matched tile--arm error ratio \(e_{\rm emu}/e_{\rm real}\) span \(0.97\)--\(0.98\), and the pre-registered O5 fidelity gate passes. The deployment test now carries this same-rank comparison to model scale.

\subsection{Model-Level Loss and the Scope of the Evidence}

Accumulated model loss is the deployment question motivated by FalconGEMM's practical use of Strassen-type fp8 kernels \citep{falcongemm}. We compare the exactly specified \(\Phi\)-optimal realization certified in Section~3 with classic Strassen at identical rank on real \texttt{deep\_gemm} blk128 kernels over thirty-two paired consecutive wikitext-2 chunks \citep{wikitext2}. Excess negative log-likelihood (NLL) is measured against the clean unquantized model, and the reported reduction quantifies what replacing classic Strassen with the \(\Phi\)-optimal realization removes. That primary reduction is \(55.0\%\) on Qwen2.5-72B-Instruct, from \(0.10810\) to \(0.04866\) nats, and \(10.2\%\) on Llama-3.3-70B-Instruct, from \(0.13529\) to \(0.12155\). The corresponding secondary KL changes are \(-47.58\%\) and \(-7.99\%\). The Qwen primary reduction and both KL comparisons are strongly significant; the Llama primary excess-NLL sign test gives \(p=0.0501\). Table~\ref{tab:llm-deploy} reports the paired win counts; Appendix~E gives the full test table. These two roughly 70B-model traces are a numerical-relevance sanity check, not a separate LLM contribution. They cover wikitext-2 only, and inference from consecutive chunks assumes exchangeability. Here, the implementations remain at \(\ell=1\) or flatten the Kronecker construction into a single nonrecursive realization.

\begin{table}[t]
\centering
{\small
\setlength{\tabcolsep}{3pt}
\begin{tabular}{@{}lcc@{}}
\toprule
Metric & Qwen2.5-72B & Llama-3.3-70B \\
\midrule
Excess-NLL reduction &
\(55.0\%\;(30/32)\) &
\(10.2\%\;(22/32)\) \\
KL change &
\(-47.58\%\;(32/32)\) &
\(-7.99\%\;(27/32)\) \\
NLL sign test & strong & \(p=0.0501\) \\
\bottomrule
\end{tabular}}
\caption{The certified \(\Phi\)-optimal realization lowers both excess NLL and KL relative to classic Strassen. Parentheses give paired wins out of 32 consecutive wikitext-2 chunks.}
\label{tab:llm-deploy}
\end{table}

What the tests support is the model's ordering, not its predicted magnitude. The prediction and real-operand tests support the \(\Phi\)-predicted ordering within the tested fused block-scaled regimes, and the deployment test supports the \(\Phi\)-optimal versus classic Strassen comparison. The model's calibration, the predicted-to-measured error ratio, spreads by up to \(2.60\%\) within an operand family across shapes and up to \(12.46\%\) across families within a shape; its \(0.37\%\) coefficient of variation across fifteen members of a separate emulated orbit sweep holds for one shape and one signed-uniform family, not as a general guarantee. A pre-registered check on that calibration fit could not separate contamination from genuine family-dependent fit quality, as detailed in Appendix~D. The ordering itself also has a boundary: population second moments alone do not determine realized error, since a family with matched second moments departs from the Gaussian family's realized error by up to \(0.189\) in relative terms. Consistent with that, on discrete Rademacher \((\pm1)\) operands classic Strassen wins all six designed synthetic cells, in the same fused block-scaled regime: what bounds the ordering is the operand distribution, not the arithmetic regime alone.

The remaining bounds concern deployment and mathematical scope. Throughput is hardware-dependent: the H20 fused kernel is free or faster than cubic fp8 in the tested shapes, while on H100 and H200 the \(\Phi\)-optimal rank-7 realization runs slower than cubic fp8, with net changes from \(-21.2\%\) to \(-11.3\%\) and from \(-15.4\%\) to \(-9.9\%\), because encoding and decoding dominate; against classic Strassen at identical rank the largest measured H100 gap is \(0.91\%\), so this is a comparison against cubic rather than a same-rank penalty. Task-level accuracy remains unresolved for lack of samples: at \(n=1000\) the paired tests give \(p=0.201\) on ARC \citep{arc-challenge} and \(p=0.888\) on HellaSwag \citep{hellaswag}, against estimated requirements of \(n\approx4{,}260\) and \(n\approx98{,}000\) to resolve the observed effects at the 5\% level with 80\% power. We make no border-rank, approximate, complex, or larger-shape optimality claim. Appendix~E gives the complete protocols, controls, and tables.

\section{Conclusion}

The result rests on three distinct links forming a single chain. An exact symbolic certificate fixes the global optimum of the coefficient functional \(\Phi\) over the \(\langle 2,2,2;7\rangle\) change-of-basis orbit; de Groote's classification extends it to the entire exact real rank-7 class as a corollary. A scoped second-moment model gives this geometry a leading-order interpretation in expected fp8 error, with realization ordering as its primary prediction. Experiments test the \(\Phi\)-predicted ordering in realized fp8 kernels, supporting it within the tested fused block-scaled regimes. The \(\Phi\)-optimal realization lies in the real-kernel low-error region. At model scale, it lowers excess NLL relative to classic Strassen for Qwen2.5-72B and Llama-3.3-70B on wikitext-2.

The links have distinct scopes. The certificate applies only to \(\Phi\)-optimality; it neither inherits the model's physical uncertainty nor supplies a lower bound on realized fp8 error. For \(\langle 4,4,4\rangle\), multiplicativity carries the \(\Phi\)-optimality certificate only to the composite \(\langle 4,4,4;49\rangle\) change-of-basis orbit of the Kronecker point, not to a global optimum. At the empirical layer, absolute magnitude is family-calibrated, the Llama comparison is marginal, and Section~4 fixes the empirical scope of the chain, including the discrete \(\pm1\) Rademacher exception. Broader real-kernel validation and device-general speed modeling remain open.

Algorithm realization is an independent low-precision design axis alongside operand rotation, with certified coefficient geometry and measured fp8 reach. The chain turns realization choice into a mathematical object: the \(\Phi\)-optimum comes from a proof, not a search.

\section*{Reproducibility Artifact}
A curated artifact will be released at
\resourcelink{\artifacturl}{\artifactdisplay}. It contains the exact certificate checkers
and certified coordinates, the fixed orbit-response panel, sanitized real-tile measurements,
and the per-chunk inputs and analysis scripts for the model-level experiments. The artifact
supports verification and reanalysis of the released data. Repeating the original GPU
measurements additionally requires third-party model checkpoints, WikiText-2, and DeepGEMM,
which are not redistributed. Author-created code, released measurements, and documentation
carry the MIT License.

\bibliographystyle{plainnat}
\bibliography{main}

\clearpage
\appendix
\section*{Technical Appendix}
The main text states the claims and their essential arguments. Appendices~A--E supply the
exact certificate arithmetic and equality cases, the full fp8 derivation, and the
reproducibility protocols with complete results.

\counterwithin{figure}{section}
\counterwithin{table}{section}
\numberwithin{equation}{section}
\renewcommand{\theHfigure}{appendix.\thesection.\arabic{figure}}
\renewcommand{\theHtable}{appendix.\thesection.\arabic{table}}
\renewcommand{\theHequation}{appendix.\thesection.\arabic{equation}}

\section{Proof of the \texorpdfstring{$\Phi$}{Phi} Certificate}
\label{app:certificate}
Section~3 sends the reader here for the coordinate expansion, full moment-map arithmetic, closed-orbit attainment, second-order form, and equality case.  Write $D=\{(U_r,V_r,W_r)\}_{r=1}^7$ and $\Phi(D)=\sum_r\|U_r\|_F^2\|V_r\|_F^2\|W_r\|_F^2$.  Throughout, a coefficient array is identified with its flattening, so the entrywise Euclidean norm and the Frobenius norm coincide and $\|\cdot\|_F$ here is the $\|\cdot\|_2$ of Section~2 --- never a spectral norm.  $\Phi$ and $\Gamma$ are therefore the same functionals the model of Section~2 and the experiments of Appendix~\ref{app:protocol} use.  With row vectorization $\operatorname{rv}(A)=(a_{11},a_{12},a_{21},a_{22})$ and the cyclic third factor $w_r=\operatorname{rv}(W_r^\top)$, the right action is
\begin{equation}
\begin{aligned}
 u_r&\mapsto u_r(X\otimes Y^{-\top}),\\
 v_r&\mapsto v_r(Y\otimes Z^{-\top}),\\
 w_r&\mapsto w_r(Z\otimes X^{-\top}).
\end{aligned}
\label{eq:a-action}
\end{equation}

\paragraph{Descent and convexity.}
For $g\in\mathrm{GL}_2(\mathbb R)^3$, use the componentwise polar decomposition $g=PQ$, not $QP$.  Under the right action, $T^*\mathbin{\cdot}g=(T^*\mathbin{\cdot}P)\mathbin{\cdot}Q$, so $Q\in\mathrm O(2)^3$ acts last and preserves $\Phi$.  Writing $P_i=s_i\widehat P_i$ with $\det\widehat P_i=1$, the central scalars act on $(U,V,W)$ by $(s_X/s_Y,s_Y/s_Z,s_Z/s_X)$, whose product is one.  Hence
\[
 F([\widehat P]):=\Phi(T^*\mathbin{\cdot}\widehat P),\qquad
 [\widehat P]\in\mathcal X:=\bigl(\mathrm{SL}_2(\mathbb R)/\mathrm{SO}_2\bigr)^3.
\]
With $v^*=(u_r\otimes v_r\otimes w_r)_{r=1}^7$ and $R$ induced by \eqref{eq:a-action}, $F=\|R(\widehat P)v^*\|^2$, twice the conventional $\kappa=2$ Kempf--Ness norm square; the factor does not affect convexity or minimizers.  For any geodesic $\sigma(t)=g_0e^{tH}K$, $H\in\mathrm{Sym}_0(2)^3$, self-adjointness of $dR(H)$ gives
\begin{equation}
\begin{aligned}
 F(\sigma(t))&=\sum_\lambda e^{2\lambda t}\|w_\lambda\|^2,\\
 F''(\sigma(t))&=4\sum_\lambda\lambda^2e^{2\lambda t}\|w_\lambda\|^2\geq0.
\end{aligned}
\label{eq:a-convexity}
\end{equation}
Thus $F$ is geodesically convex on the Hadamard manifold $\mathcal X$, the real-group form of Kempf--Ness convexity \citep{bfgoww-2019}.

\paragraph{Coordinates and moment map.}
Under our right-action convention, the three equal sandwich transforms use the inverse of the minimizing DPS Proposition~28 factor \citep{dumas-pernet-sedoglavic-2025}:
\[
\begin{gathered}
 T_\rho=\begin{bmatrix}\rho^{-1}&\rho/2\\0&\rho\end{bmatrix},
 \qquad \rho=(4/3)^{1/4},\\[1mm]
 K_\rho=T_\rho\otimes T_\rho^{-\top}
 =\begin{bmatrix}
 1&0&\rho^2/2&0\\
 -1/2&\rho^{-2}&-\rho^2/4&1/2\\
 0&0&\rho^2&0\\
 0&0&-\rho^2/2&1
 \end{bmatrix}.
\end{gathered}
\]
They are transforms; $T^*$ is the decomposition.  Put $\xi=1/\sqrt3$, $\eta=\sqrt3/2$, $\zeta=1/(2\sqrt3)$.  Applying $K_\rho$ to the cyclic Strassen arrays gives $U^*=[u_1;\ldots;u_7]$ with
\begin{equation}
\begin{gathered}
 U^*=\begin{bmatrix}
1&0&0&1\\0&0&\xi&1\\1&0&\xi&0\\0&0&-\xi&1\\
1/2&\eta&\zeta&1/2\\-1&0&\xi&0\\-1/2&\eta&\zeta&-1/2
\end{bmatrix},\\[1mm]
 V^*_{r,:}=U^*_{\pi_V(r),:},\qquad
 w^*_r=U^*_{\pi_W(r),:}.
\end{gathered}
\label{eq:a-coordinates}
\end{equation}
Here $\pi_V=(1,3,7,6,4,5,2)$ and $\pi_W=(1,7,2,5,6,4,3)$.  The third relation is stated in the $w$ coordinates fixed above, so as a coefficient array $W^*_r$ is the transpose of the $\pi_W(r)$th block of $U^*$, not a copy of it.  Thus the three coefficient arrays are distinct even though the sandwich transforms are equal.  Their squared row norms coincide:
\begin{equation}
\begin{aligned}
 a_r&:=\|U_r^*\|_F^2=\|V_r^*\|_F^2=\|W_r^*\|_F^2,\\
 (a_r)_{r=1}^7&=(2,4/3,4/3,4/3,4/3,4/3,4/3).
\end{aligned}
\label{eq:a-norms}
\end{equation}
Consequently, $\Phi(T^*)=2^3+6(4/3)^3=200/9$.

Reshape each row as a $2\times2$ matrix and set $\mathcal L_A=\sum_ra_r^2A_rA_r^\top$, $\mathcal R_A=\sum_ra_r^2A_r^\top A_r$.  Put
$B_\pm=\left[\begin{smallmatrix}32/9&\pm32\sqrt3/27\\\pm32\sqrt3/27&32/27\end{smallmatrix}\right]$ and
$C_\pm=\left[\begin{smallmatrix}32/27&\pm32\sqrt3/27\\\pm32\sqrt3/27&32/9\end{smallmatrix}\right]$.
The full arithmetic is
\begin{align}
\mathcal L_U
 &=4I_2+\begin{bmatrix}0&0\\0&128/27\end{bmatrix}+B_++B_-
 =\frac{100}{9}I_2,
\label{eq:a-left}\\
\mathcal R_U
 &=4I_2+\begin{bmatrix}128/27&0\\0&0\end{bmatrix}+C_++C_-
 =\frac{100}{9}I_2.
\label{eq:a-right}
\end{align}
In \eqref{eq:a-left}, the displayed terms group rows $(1)$, $(2,4)$, $(3,5)$, $(6,7)$; in \eqref{eq:a-right}, they group $(1)$, $(3,6)$, $(2,5)$, $(4,7)$.  The permutations in \eqref{eq:a-coordinates} preserve the exceptional first weight, hence
\begin{equation}
 \mathcal L_A=\mathcal R_A=\frac{100}{9}I_2
 \quad(A=U^*,V^*,W^*).
\label{eq:a-isotropy}
\end{equation}
For $H_X,H_Y,H_Z\in\mathrm{Sym}_0(2)$,
\[
\begin{aligned}
 \frac12F'(0)
 &=\operatorname{tr}H_X(\mathcal L_U-\mathcal R_W)\\
 &\quad+\operatorname{tr}H_Y(\mathcal L_V-\mathcal R_U)\\
 &\quad+\operatorname{tr}H_Z(\mathcal L_W-\mathcal R_V).
\end{aligned}
\]
These three traceless brackets are the moment-map components, and \eqref{eq:a-isotropy} gives the exact symbolic identity
\begin{equation}
 \mu(T^*)=(0,0,0),\qquad \nabla F([I])=0.
\label{eq:a-moment-zero}
\end{equation}
Equations~\eqref{eq:a-convexity} and \eqref{eq:a-moment-zero} prove global minimality.  The real Kempf--Ness minimal-vector criterion also makes the orbit closed, so $T^*$ is polystable and the minimum is attained \citep{richardson-slodowy-1990}.  Attainment is not automatic for semistability: under $s\cdot(x_+,x_0,x_-)=(sx_+,x_0,s^{-1}x_-)$, the semistable point $(1,1,0)$ has nonclosed orbit and norm-square $1+s^2$, whose infimum $1$ occurs only at the closure point $(0,1,0)$.

\paragraph{Second order and equality.}
Let $H(p,q)=\tfrac12\left[\begin{smallmatrix}p&q\\q&-p\end{smallmatrix}\right]$ and $t=(x_1,y_1,z_1,x_2,y_2,z_2)^\top$, with $(H_X,H_Y,H_Z)=(H(x_1,x_2),H(y_1,y_2),H(z_1,z_2))$.  Set $\Phi_t=\Phi(T^*\cdot(e^{H_X},e^{H_Y},e^{H_Z}))$.  Exact expansion gives
\begin{equation}
\begin{aligned}
 \Phi_t&=\frac{200}{9}\\
 &\quad+\frac89t^\top\operatorname{diag}(Q,Q)t+O(\|t\|^3),\\
 Q&=\left[\begin{smallmatrix}
 21&-13/2&-13/2\\-13/2&21&-13/2\\-13/2&-13/2&21
 \end{smallmatrix}\right].
\end{aligned}
\label{eq:a-hessian}
\end{equation}
The eigenvalues of $Q$ are $8$ once and $55/2$ twice; across both blocks they have multiplicities $2$ and $4$.  The coordinate Hessian eigenvalues are instead $128/9$ with multiplicity $2$ and $440/9$ with multiplicity $4$.  These six coordinates are all of $\mathrm{Sym}_0(2)^3$.

The equality case unwinds in five steps.  \emph{(1)} Section~\ref{app:degroote} writes any exact real rank-$7$ $D$, modulo permutation and product-one termwise rescaling, as $T^*\cdot g$.  \emph{(2)} The $PQ$ reduction gives $\Phi(D)=F(x)$ for the unique determinant-one SPD representative $x\in\mathcal X$, with the trailing orthogonal factor removed.  \emph{(3)} Convexity and \eqref{eq:a-moment-zero} give $F\geq200/9$, while \eqref{eq:a-hessian} is positive definite.  \emph{(4)} If $y\neq[I]$ were another minimizer, the unique Hadamard geodesic from $[I]$ to $y$ would have nonzero initial tangent; endpoint minimality and convexity would make $F$ constant on it, forcing second derivative zero and contradicting \eqref{eq:a-hessian}.  \emph{(5)} Hence the SPD representative is $I$.  The remaining central factors act by $(s_X/s_Y,s_Y/s_Z,s_Z/s_X)$ and are an admissible product-one rescaling; the remaining $Q$ is orthogonal.  Equality therefore holds exactly for an orthogonal sandwich, term permutation, and admissible product-one termwise rescaling of $T^*$.  The absorbed discrete stabilizer is already represented by those permutation and rescaling gauges and adds no equality class.

The accompanying artifact includes \texttt{certificate/}\allowbreak\texttt{check\_floor.py}, which derives the exact stationarity identity, Hessian matrix, and spectrum symbolically over $\mathbb Q(\rho)$ and, under \texttt{verify.py -{}-control}, rejects a deliberately perturbed algebraic point.  Geodesic convexity and Hadamard uniqueness are established analytically; the sampled-geodesic checks are numerical controls, not proof components.  The random-restart objective gap \texttt{-1.78e-14} and the sampled-geodesic nonnegative-exponential fit residual \texttt{1.58e-15} are separate controls; \eqref{eq:a-moment-zero} is the exact moment-map result.

\section{The de Groote Rank-\texorpdfstring{$7$}{7} Reductions}
\label{app:degroote}
Section~3 sends the reader here for the theorem and reductions that lift the orbit result to the exact real rank-$7$ class.  For $D=\{(U_r,V_r,W_r)\}_{r=1}^7$, put $t_r=U_r\otimes V_r\otimes W_r$ and $\Phi(D)=\sum_r\|t_r\|_{\mathrm{HS}}^2$.

De Groote's Theorem~0.1(i) states over an arbitrary field $K$ that every exact rank-$7$ decomposition of $\langle2,2,2\rangle_K$ is obtained from Strassen's $\gamma$ by a $K$-rational sandwich $g\in\mathrm{GL}_2(K)^3$, a permutation $\pi\in\mathfrak S_7$, and nonzero term scalars $(\alpha_r,\beta_r,\gamma_r)$ with $\alpha_r\beta_r\gamma_r=1$, the symbols Section~2 uses for this rescaling \citep{degroote-1978}.  Taking $K=\mathbb R$ supplies the required real construction.  Burichenko corroborates the arbitrary-field transitivity statement while citing de Groote, but does not reprove it \citep{burichenko-2015}.  Chiantini et al. alone do not supply the real result: their transitivity theorem is over $\mathbb C$ \citep{cilo-2019}, and one complex orbit may contain more than one real orbit.

The invariance legs are as follows.  \emph{DG1:} $\|t_r\|_{\mathrm{HS}}^2=\|U_r\|_F^2\|V_r\|_F^2\|W_r\|_F^2$, so $\Phi$ depends on the seven rank-one tensors, not their factorization.  \emph{DG2:} permutations reorder the sum, and product-one rescaling leaves each $t_r$ unchanged; the sampled worst relative drift is \texttt{6.4e-16}.  \emph{DG3:} a trailing orthogonal sandwich, including determinant-$-1$ components, acts by orthogonal Kronecker maps; the sampled worst relative drift is \texttt{1.1e-15}.  \emph{DG3b:} central factors scale $(U,V,W)$ by $(s_X/s_Y,s_Y/s_Z,s_Z/s_X)$, whose product is exactly one; its floating-point test is a sampled sanity check of this displayed $s^0$ cancellation.  These legs are analytically invariant, with sampled numerical checks at the reported residuals; the samples do not prove group completeness or universal invariance.  The discrete stabilizer within the factor-preserving sandwich action is isomorphic to $S_3$ and acts through the listed permutation and rescaling gauges, adding neither a separate gauge nor another equality class.  A general nonorthogonal sandwich can change $\Phi$ and is the continuous orbit freedom minimized in Section~\ref{app:certificate}.

As a disconfirming sanity control only, DG4 optimized all $84$ free coefficients of seven unrestricted rank-one triples from $400$ starts, without assuming Strassen's orbit.  Of the $206$ exactness-valid endpoints, the smallest value was $22.226466>200/9$: the search approached the analytic floor from above and did not breach it.  The proof remains de Groote's classification, the analytic DG1--DG3b identities, and Section~\ref{app:certificate}.

\section{The \texorpdfstring{$\Gamma$}{Gamma} Growth-Factor Argument}
\label{app:gamma}
Section~3 sends the reader here for the DPS-gap globality argument and the executable-witness disclosure.  Define $\lambda_r=\|U_r\|_F\|V_r\|_F\|W_r\|_F$ and $\Gamma(D)=\sum_r\lambda_r=\|\lambda\|_1$, while $\Phi=\|\lambda\|_2^2$.

At the $\Phi$-optimal realization $T^*$, the squared factor norms are $a_1=2$ and $a_r=4/3$ for $r=2,\ldots,7$, so
\begin{equation}
 \Gamma_{\min}=\Gamma(T^*)
 =2\sqrt2+6\frac{8}{3\sqrt3}
 =2\sqrt2+\frac{16}{\sqrt3}.
\label{eq:c-gamma-value}
\end{equation}
The cyclic coefficient matrices also satisfy, for $A\in\{U^*,V^*,W^*\}$,
\begin{equation}
 \sum_r\sqrt{a_r}\,A_rA_r^\top
 =\sum_r\sqrt{a_r}\,A_r^\top A_r
 =\left(\sqrt2+\frac8{\sqrt3}\right)I_2.
\label{eq:c-gamma-stationarity}
\end{equation}
Indeed, the $X$ derivative of one summand is $\sqrt{a_r}\operatorname{tr}H_X(U_rU_r^\top-W_r^\top W_r)$, with cyclic analogues for $Y,Z$; hence \eqref{eq:c-gamma-stationarity} gives exact criticality.

For any geodesic $g_0e^{tH}K$ and any factor $z_{rj}$ of term $r$, self-adjointness gives
\[
\begin{aligned}
 S_{rj}(t)&:=\|z_{rj}(t)\|^2
 =\sum_\omega c_{rj,\omega}e^{2\omega t},\qquad c_{rj,\omega}\geq0,\\
 p_{rj,\omega}(t)&=\frac{c_{rj,\omega}e^{2\omega t}}{S_{rj}(t)}.
\end{aligned}
\]
Thus $(\log\|z_{rj}\|)''=2\operatorname{Var}_{p_{rj,t}}(\omega)$.  For $f_r=\|u_r\|\|v_r\|\|w_r\|$, let $\operatorname{Var}_r$ be the sum of the three variances.  Then
\begin{equation}
\begin{aligned}
 (\log f_r)''&=2\operatorname{Var}_r\geq0,\\
 f_r''&=f_r\bigl((\log f_r)''+((\log f_r)')^2\bigr)\geq0.
\end{aligned}
\label{eq:c-summand-convexity}
\end{equation}
Every summand, and hence $\Gamma$, is geodesically convex.  Exact criticality therefore gives the global minimum \eqref{eq:c-gamma-value}, as convexity makes every critical point global \citep{bfgoww-2019}; the de Groote reduction lifts it to the entire exact real rank-$7$ class because permutation, product-one rescaling, and orthogonal sandwiches preserve $\Gamma$.  This closes the $\leq2.6\%$ gap stated by DPS without a restart argument \citep{dumas-pernet-sedoglavic-2025}.

Equations \eqref{eq:c-gamma-stationarity} and \eqref{eq:c-summand-convexity} give the complete analytic globality proof; the exactness-gated computation independently verifies the value $12.066031431780$.  The sampled-geodesic check, which found zero violations among $72$ accepted of $120$ attempted, and the separate $80$-restart optimization, which agreed to floating-point precision, are numerical sanity controls rather than proof components.

\section{The Second-Moment Model Derivation}
\label{app:model}

The final sentence of Section~2 points here for the complete calculation behind the leading-order expected-error model and for the calibration-transport check.  Let full matrices \(A\in\mathbb{R}^{M\times K}\) and \(B\in\mathbb{R}^{K\times N}\) be partitioned according to a bilinear algorithm \(\langle m,k,n;R\rangle\), with within-block dimensions \(P=M/m\), \(L=K/k\), and \(Q=N/n\).  For realization
\(D=\{(U_r,V_r,W_r)\}_{r=1}^R\), write
\[
\Phi(D)=\sum_{r=1}^R\|U_r\|_2^2\|V_r\|_2^2\|W_r\|_2^2.
\]

\subsection{Noise assumptions and expansion}

At a fixed within-block output position, term \(r\) forms encoded inner-coordinate scalars
\(x_{r\ell}=\langle U_r,a_\ell\rangle\) and
\(y_{r\ell}=\langle V_r,b_\ell\rangle\), \(1\leq \ell\leq L\).  The original operand blocks remain unquantized; fused blk128 e4m3 quantization is applied only after these encoded combinations have been formed.  The operand entries are iid and zero mean on each side, the two operand arrays are independent, and their variances are \(\sigma_A^2\) and \(\sigma_B^2\).  The rounding errors \(\eta^A_{r\ell}\) and \(\eta^B_{r\ell}\) are zero mean and independent of the operands; their cross moments vanish across operand side, term \(r\), and inner coordinate \(\ell\), and the two side-noise squares factor for the product term.  Their normalized second moments are
\begin{equation}
\begin{aligned}
\mathbb{E}x_{r\ell}^2&=\|U_r\|_2^2\sigma_A^2,
&\mathbb{E}(\eta^A_{r\ell})^2&=s_A^2\|U_r\|_2^2\sigma_A^2,\\
\mathbb{E}y_{r\ell}^2&=\|V_r\|_2^2\sigma_B^2,
&\mathbb{E}(\eta^B_{r\ell})^2&=s_B^2\|V_r\|_2^2\sigma_B^2.
\end{aligned}
\label{eq:model-noise-moments}
\end{equation}
The product error accumulated by term \(r\) at that output position is
\begin{equation}
 t_r=\sum_{\ell=1}^{L}
 \left(\eta^A_{r\ell}y_{r\ell}
       +x_{r\ell}\eta^B_{r\ell}
       +\eta^A_{r\ell}\eta^B_{r\ell}\right).
\label{eq:model-term-error}
\end{equation}
Independence from the operands removes the mixed operand--noise moments, side decorrelation removes the cross terms within one summand, and inner-coordinate decorrelation removes the \(\ell\neq\ell'\) terms.  Hence
\begin{align}
\mathbb{E}t_r^2
&=L\left(s_A^2+s_B^2+s_A^2s_B^2\right)
  \|U_r\|_2^2\|V_r\|_2^2\sigma_A^2\sigma_B^2.
\label{eq:model-one-term}
\end{align}
In the termwise gauge \(\|U_r\|_2=\|V_r\|_2=1\), this is the canonical scalar expression
\(L(s_A^2+s_B^2+s_A^2s_B^2)\sigma_A^2\sigma_B^2\); restoring the gauge gives Eq.~\eqref{eq:model-one-term}.

The decoder sends this scalar error to the \(mn\) output blocks as \(t_rW_r\).  Cross-term decorrelation across \(r\) therefore gives, for one within-block output position,
\[
\mathbb{E}\left\|\sum_{r=1}^R t_rW_r\right\|_2^2
=L\left(s_A^2+s_B^2+s_A^2s_B^2\right)
 \sigma_A^2\sigma_B^2\Phi(D).
\]
There are \(PQ\) such positions.  Since \(PQL=MNK/(mkn)\),
\begin{equation}
\mathbb{E}\|\delta C\|_F^2
=\frac{MNK}{mkn}
 \left(s_A^2+s_B^2+s_A^2s_B^2\right)
 \sigma_A^2\sigma_B^2\Phi(D).
\label{eq:model-numerator}
\end{equation}
For iid zero-mean operands, each exact output entry is a sum of \(K\) independent products, so
\begin{equation}
\mathbb{E}\|C\|_F^2=MNK\,\sigma_A^2\sigma_B^2.
\label{eq:model-denominator}
\end{equation}
Dividing Eqs.~\eqref{eq:model-numerator} and \eqref{eq:model-denominator} yields
\begin{equation}
\frac{\mathbb{E}\|\delta C\|_F^2}{\mathbb{E}\|C\|_F^2}
=\frac{\left(s_A^2+s_B^2+s_A^2s_B^2\right)\Phi(D)}{mkn}.
\label{eq:model-full-second-moment}
\end{equation}
With \(s_A=s_B=s_q\), retaining only the leading operand-side terms gives
\begin{equation}
\frac{\mathbb{E}\|\delta C\|_F^2}{\mathbb{E}\|C\|_F^2}
\approx \frac{2s_q^2\Phi(D)}{mkn},
\qquad
\mathrm{err}_{\mathrm{RMS}}\approx s_q\sqrt{\frac{2\Phi(D)}{mkn}}.
\label{eq:model-leading}
\end{equation}
Relating this ratio of expectations to a realized relative error additionally assumes concentration.  Likewise, the disappearance of cross-\(r\) covariance is an assumption rather than an algebraic identity; without it, the numerator contains
\(\sum_{r,s}\langle W_r,W_s\rangle\operatorname{Cov}(t_r,t_s)\) instead of the diagonal sum defining \(\Phi\).

We measure that discarded term rather than only naming it.  Decomposing the emulated error into its part \(L\) that is linear in the rounding residuals and comparing \(\lVert L\rVert_F^2\) against the diagonal sum \(\sum_{r,p}W_{rp}^2\lVert t_r\rVert_F^2\) that defines \(\Phi\) gives the residual fraction \((\lVert L\rVert_F^2-\mathrm{diag})/\lVert L\rVert_F^2\).  In a six-arm diagnostic at \(n=512\) on iid Gaussian operands with four seeds, the armwise mean of that fraction ranges from \(-0.085\%\) to \(0.201\%\) under the emulated \(1\times128\) block-scaled convention.  Three limits belong with those figures.  They are armwise means over the four seeds rather than per-seed extrema, so they describe the term rather than bound it; they cover the linear component alone, not the second-order or mixed contributions; and the diagnostic groups both operands in \(1\times128\), where the kernel groups weights in \(128\times128\).  What they establish is that the realized cross-\(r\) contribution is small in this diagnostic, not that the covariance parameter itself is bounded, and not that Assumption~1 holds in general.

Here \(s_q\) is the per-block operand-rounding noise coefficient for the encoded blocks, measured separately under the selected operand-side reference and held fixed across algorithm realizations.  It is distinct from the runtime group scale \(\operatorname{amax}/448\).  The initial diagnostic reference rounded the product; after inspecting its fit, we switched post hoc to the operand-side reference used in the paper and interpret \(s_q\) throughout as operand-side rounding in \(x_r\) and \(y_r\).  This choice precludes an absolute-calibration claim; accordingly, ordering rather than magnitude is the model's primary output.  Keeping the product term in Eq.~\eqref{eq:model-term-error} changes \(2s_q^2\) to \(2s_q^2+s_q^4\).  Its relative contribution is \(s_q^2/2\approx3.2\times10^{-4}\), against a measured \(3.3\times10^{-4}\), so dropping it has negligible numerical effect at the calibrated noise level.

\paragraph{Relation to quantized-matmul rate--distortion theory.}
Section~1 groups this neighbouring line of work with the operand-side axis; the distinction is worth stating precisely here, since both predict an expected error for a quantized product.  Ordentlich and Polyanskiy characterize the information-theoretic distortion of a quantized product at a given bit rate, including for the calibration-free setting in which both factors are quantized, and compare absmax INT and floating-point schemes against that limit \citep{ordentlich-2026, ordentlich-highrate-2026}.  Their free variable is the quantizer: the product computed is the standard one, and rate is spent to minimize distortion.  Ours is the algorithm: the quantizer is fixed at block-scaled e4m3 with no rate budget to allocate, and the freedom is the change-of-basis orbit of an exact rank-\(R\) decomposition.  The two are composable rather than competing --- their bound constrains what any quantizer can achieve at a given rate, while \(\Phi\) constrains how much a realization amplifies whatever operand noise the quantizer leaves.  Neither quantity bounds the other: their limit is a rate-indexed average-case MSE for the standard product, and \(\Phi_{\min}=200/9\) is a fixed-precision coefficient floor across exact rank-7 realizations.

\subsection{Calibration transport and retirement of the contamination check}

The pre-registered contamination check calibrated \(s_q\) on a source operand family, froze that calibration, and transported the predicted-to-measured ratio
\begin{equation}
\mathcal{R}(F,S,D)=
\frac{s_q\sqrt{2\Phi(D)/(mkn)}}{e_{\mathrm{meas}}(F,S,D)}
\label{eq:model-calibration-ratio}
\end{equation}
to held-out operand families \(F\), shapes \(S\), and realizations \(D\).  The differential-fit criterion failed in the original run and failed again with a neutral-calibration corpus.  The follow-up therefore retired this check: comparing source and held-out fit cannot separate contamination from genuine family-dependent fit quality.

The magnitude summaries have separate provenances.  The original, separate cross-family calibration experiment produced the earlier \(\pm15\%\) family-conditional band.  An independent reproduction on three new shapes used the spread \(100(\max \mathcal{R}-\min \mathcal{R})/\operatorname{mean}\mathcal{R}\): its maximum within-family, across-shape spread was \(2.60\%\), and its maximum within-shape, across-family spread was \(12.46\%\).  The \(12.46\%\) reproduction is not a rounding of the earlier \(\pm15\%\) band.  Both measurements are spreads of \(\mathcal{R}\) rather than its displacement from one, so what they establish is that the ratio transports consistently within a family and less so across families; a predictor carrying a constant multiplicative bias would show the same small spread. They are therefore evidence that magnitude is family-conditional, not that its absolute level is calibrated. Eq.~\eqref{eq:model-leading}'s ordering prediction remains the primary model output.

\section{Experimental Protocol and Full Results}
\label{app:protocol}

The final sentence of Section~4 points here for the complete protocols, controls, and tables.  For local reference, a realization \(D=\{(U_r,V_r,W_r)\}_{r=1}^R\) has
\(\Phi(D)=\sum_r\|U_r\|_2^2\|V_r\|_2^2\|W_r\|_2^2\) and
\(\Gamma(D)=\sum_r\|U_r\|_2\|V_r\|_2\|W_r\|_2\).  The model-consistent cross-shape predictor is \(\sqrt{\Phi/(mkn)}\).

\subsection{Computing environment}

Two platforms carry the experiments.  The real-tile panel of Section~4.2 and the two-model deployment test of Section~4.3 ran on an on-premises node with eight NVIDIA H20 accelerators of \(96\)\,GB each, two 96-core AMD EPYC server processors giving \(384\) logical cores, and roughly \(2.3\)\,TB of host memory, running a vendor-maintained enterprise Linux distribution with kernel 5.4.241, against CUDA 12.8.  The deployment runs of Section~4.3 record PyTorch 2.10.0; the panel runs of Section~4.2 predate that echo and record no version.  The cross-hardware throughput comparison ran on NVIDIA H100 and H200 allocations rented from a commercial GPU cloud, under PyTorch 2.11.0 with Triton 3.6.0 against CUDA 12.8.1; the reported H100 allocations report the device string \texttt{NVIDIA H100 80GB HBM3}, and because that provider's H100 pool mixes SXM and NVL variants, the device string is recorded per run rather than assumed constant.  The fp32 emulator is device-independent.  Host hardware and operating system for the on-premises node are as observed on that node.

The \texttt{deep\_gemm} build is pinned unevenly across these platforms, and we state that rather than smooth it.  On the rented allocations the kernel was cloned from its default branch and its commit hash echoed into the log from the second round of those runs onward; every reported H100 and H200 run records \texttt{559d79fb6994a58b8a15}\allowbreak\texttt{b4b93bf13ccc16edf247}, while the earlier scout allocation predates that echo and its commit is unlogged.  On the H20 node the kernel was imported from the system installation and its upstream revision was not recorded when the runs were made; re-examined on that node, the install carries no version attribute, no packaging metadata and no repository metadata, and the compiled object embeds no commit string, so no upstream revision is recoverable.  The binary identity was recorded as the SHA-256 of the compiled \texttt{deep\_gemm} shared object together with its installation date.  These machine-specific records are not part of the artifact, so we do not claim byte-level regeneration of the original kernel measurements.  This limitation does not affect the quantization semantics stated below, which are fixed by our harness rather than by the kernel build.  The emulator reproduces the e4m3 scaling and rounding rule exactly; it differs from the real path in weight-scale granularity and in the bf16 pre-round, as the next paragraph sets out.

\subsection{Emulation, real-kernel sampling, and the twenty-shape sweep}

The breadth experiment used an fp32 emulator of fused blk128 e4m3: raw operands remained unquantized, encoded combinations were block-scaled and rounded, and relative Frobenius error was measured against an fp64 product.  Each AlphaTensor realization \citep{alphatensor} used one level of fast multiplication, one signed-uniform synthetic operand family, eight paired samples at block edge 256, and identical operands and seeds before and after re-basing.  \emph{Re-basing} a decomposition means numerically minimizing \(\Phi\) over its own change-of-basis orbit and keeping the minimizing representative: from the library point and \(23\) perturbed starts (\(X,Y,Z\) drawn as \(I\) plus Gaussian noise of scale \(0.45\), rejecting any start with \(\lvert\det\rvert<10^{-3}\)), each start is driven by L-BFGS with strong-Wolfe line search in double precision, and the lowest finite exactness-passing terminal \(\Phi\) is taken.  It adds no bilinear multiplications and leaves the shape and multiplication count \(R\) fixed.  In Table~\ref{tab:crossshape-full}, \(e\) is the mean relative Frobenius error at the library realization; \(e_{\rm lib}/e_{\rm cub}\) and \(e_{\rm reb}/e_{\rm cub}\) are the paired before- and after-re-basing errors normalized by cubic fp8.

The real-kernel experiments used the open-source fused blk128 \texttt{deep\_gemm} FP8 GEMM kernel.\footnote{\url{https://github.com/deepseek-ai/DeepGEMM}}  It has no accompanying publication; its known revision information is summarized above.  What we can state exactly is the quantization our harness applies: a block's scale is its absolute maximum divided by the e4m3 maximum \(448\) and kept in fp32, and the scaled block is cast to e4m3 under round-to-nearest-even and rescaled.  The real kernel takes \(1\times128\) groups along the contracted axis on the activation side and \(128\times128\) blocks on the weight side; the fp32 emulator described above takes \(1\times128\) groups on both.  Emulated and real runs therefore differ in the kernel and in weight-scale granularity, not in the e4m3 rounding mode; the real path additionally applies a bf16 pre-round to each encoded weight combination.  A real-operand observation is one activation-by-weight matmul for a single linear layer and context, using the first \(M\leq512\) activation rows from fixed wikitext-2 contexts \citep{wikitext2}, with \(K\) and \(N\) divisible by 256.  Errors are per-matmul against fp64, and equivalent rank-7 realizations are compared in the canonical rebalanced gauge.  That gauge is the admissible product-one termwise rescaling \((U_r,V_r,W_r)\mapsto(U_r/s_r,\,s_rV_r,\,W_r)\) with \(s_r=(\lVert U_r\rVert_2/\lVert V_r\rVert_2)^{1/2}\), which equalizes the two encoder norms, \(\lVert U_r\rVert_2=\lVert V_r\rVert_2=(\lVert U_r\rVert_2\lVert V_r\rVert_2)^{1/2}\), in every term and leaves the decoder untouched.  Fixing it is a measurement requirement rather than a cosmetic one: \(\Phi\) and \(\Gamma\) are invariant under this rescaling because they depend only on the products \(\lambda_r\), but realized fp8 error is not, since a sufficiently extreme split drives one encoded operand toward the e4m3 subnormal range and would register as accuracy that the realization does not have.  How extreme, and by what route, is worth stating precisely, because the answer bounds how much the gauge convention can be doing.  A per-block dynamic scale is positively homogeneous --- \(s(cx)=\lvert c\rvert s(x)\), so \(Q(cx)=cQ(x)\) and the e4m3 codes do not move at all --- and the rescaling above divides one encoded operand of a term by \(s_r\) while multiplying the other by it.  The two cancel term by term, and the subnormal mechanism is therefore reached not through the split as such but through the quantizer's minimum-scale clamp, once a factor drives a block's absolute maximum below \(10^{-12}\).  Measured on the emulator, the identity holds to \(6.2\times10^{-8}\) for factors down to \(10^{-12}\) and breaks to \(3.7\times10^{-2}\) at \(10^{-13}\), the first sampled factor at which the clamp engages.  On the real \texttt{deep\_gemm} path the same test returns a small but nonzero answer, and the reason is a step the emulator does not have.  Our real-path harness rounds each encoded weight combination to bf16 before its \(128\times128\) cast, and bf16 rounding is exact under scaling by a power of two but not otherwise; a power-of-two gauge accordingly leaves the measured error unchanged to the last bit, while a general gauge moves it by at most \(0.19\%\).  We read that contrast as a bound and not as an attribution: the fp32 emulator, which applies no bf16 pre-round, also moves slightly under a general gauge, so this comparison does not establish the pre-round as the only step that fails to be homogeneous.  Both figures are for the certified optimum and classic Strassen; the Winograd arm is excluded here for the reason given in Section~\ref{app:winograd-chain}.  So the canonical gauge is fixed for reproducibility and for the emulator's fixed-scale diagnostics, which do not enjoy this invariance, rather than because the panel's own numbers are materially sensitive to it over the gauges tested.  \emph{Tile} below is a synonym for one such observation; the per-tile diagnostics are computed over the same population as Eq.~\eqref{eq:real-panel-energy}.  The sampling produced \(2352\) observations across five models from four architecture families; the fixed-panel details appear below.

\begin{table*}[!t]
\centering
{\small
\setlength{\tabcolsep}{2.6pt}
\begin{tabular}{@{}lrrrrrrrr@{}}
\toprule
Shape & \(R\) & \(mkn\) & \(\Phi_{\rm lib}\) & \(x\) & \(e\) & \(e_{\rm lib}/e_{\rm cub}\) & \(e_{\rm reb}/e_{\rm cub}\) & \(e_{\rm lib}/e_{\rm reb}\) \\
\midrule
$\langle 2,2,3\rangle$ & 11 & 12 & 42 & 1.8708 & 0.0680 & 1.9046 & 1.6670 & 1.1425 \\
$\langle 2,2,2\rangle$ & 7 & 8 & 46 & 2.3979 & 0.0874 & 2.4486 & 1.7190 & 1.4244 \\
$\langle 2,2,4\rangle$ & 14 & 16 & 98 & 2.4749 & 0.0856 & 2.4003 & 1.7151 & 1.3996 \\
$\langle 2,3,3\rangle$ & 15 & 18 & 114 & 2.5166 & 0.0917 & 2.5702 & 2.0866 & 1.2317 \\
$\langle 2,2,5\rangle$ & 18 & 20 & 140 & 2.6458 & 0.0969 & 2.7139 & 2.0526 & 1.3222 \\
$\langle 2,3,4\rangle$ & 20 & 24 & 200 & 2.8868 & 0.1033 & 2.8935 & 2.2097 & 1.3094 \\
$\langle 2,4,4\rangle$ & 26 & 32 & 278 & 2.9475 & 0.1076 & 3.0145 & 2.3897 & 1.2615 \\
$\langle 2,3,5\rangle$ & 25 & 30 & 278 & 3.0441 & 0.1107 & 3.1019 & 2.2051 & 1.4067 \\
$\langle 3,4,4\rangle$ & 38 & 48 & 456 & 3.0822 & 0.1122 & 3.1441 & 2.6242 & 1.1981 \\
$\langle 3,3,4\rangle$ & 29 & 36 & 384 & 3.2660 & 0.1190 & 3.3369 & 2.4637 & 1.3545 \\
$\langle 3,3,3\rangle$ & 23 & 27 & 313 & 3.4048 & 0.1246 & 3.4934 & 2.6796 & 1.3037 \\
$\langle 3,5,5\rangle$ & 58 & 75 & 923 & 3.5081 & 0.1274 & 3.5712 & 2.7824 & 1.2835 \\
$\langle 3,4,5\rangle$ & 47 & 60 & 780 & 3.6056 & 0.1302 & 3.6509 & 2.8810 & 1.2673 \\
$\langle 3,3,5\rangle$ & 36 & 45 & 625 & 3.7268 & 0.1272 & 3.5680 & 2.4626 & 1.4488 \\
$\langle 5,5,5\rangle$ & 98 & 125 & 1785 & 3.7789 & 0.1379 & 3.8641 & 2.9117 & 1.3271 \\
$\langle 4,5,5\rangle$ & 76 & 100 & 1460 & 3.8210 & 0.1394 & 3.9098 & 3.1944 & 1.2240 \\
$\langle 2,4,5\rangle$ & 33 & 40 & 610 & 3.9051 & 0.1426 & 3.9948 & 2.5914 & 1.5415 \\
$\langle 2,5,5\rangle$ & 40 & 50 & 856 & 4.1376 & 0.1509 & 4.2348 & 2.5152 & 1.6837 \\
$\langle 4,4,5\rangle$ & 63 & 80 & 1506 & 4.3388 & 0.1583 & 4.4342 & 3.0501 & 1.4538 \\
$\langle 4,4,4\rangle$ & 49 & 64 & 3016 & 6.8648 & 0.2507 & 7.0213 & 3.5098 & 2.0005 \\
\bottomrule
\end{tabular}}
\caption{Twenty-shape fp32-emulated fused-blk128 results.  \(\Phi_{\rm lib}\) is evaluated at the library realization, \(x=\sqrt{\Phi_{\rm lib}/(mkn)}\), \(e\) is mean relative Frobenius error, and the final three columns are paired library, re-based, and reduction values from the intervention run.  Population: one signed-uniform family, eight paired samples per shape, block edge 256.}
\label{tab:crossshape-full}
\end{table*}

All Spearman calculations use average ranks for ties.  Over the twenty rows, measured \(e\) has
\(\rho_S=0.9940\) with \(\sqrt{\Phi_{\rm lib}/(mkn)}\), compared with
\(\rho_S=0.9759\) for the \(GQ_\chi/\sqrt{k}\) heuristic we adapt from the DPS coefficient-growth factor \citep{dumas-pernet-sedoglavic-2025} --- an \(\ell_2\) coefficient-growth factor that weights each term's decoder energy by whether its encoder sides require a fresh block-scaled quantization, divided by \(\sqrt{k}\) --- and
\(\rho_S=0.9455\) for bare \(\sqrt{\Phi_{\rm lib}}\).  Concretely, rescale each term to \(\lVert U_r\rVert_2=\lVert V_r\rVert_2=1\), pushing the magnitude into \(W_r\); set \(\chi^A_r=0\) if the rescaled \(U_r\) has exactly one entry of magnitude at least \(10^{-6}\), the implementation's zero threshold, and \(\chi^A_r=1\) otherwise, and \(\chi^B_r\) likewise from \(V_r\).  Then \(GQ_\chi=(\operatorname{mean}_q\sum_r W_{rq}^2(\chi^A_r+\chi^B_r))^{1/2}\), the singleton exemption reflecting that a one-entry encoder row folds into the per-product scale and so needs no fresh block scale.  The improvement from \(0.9455\) to \(0.9940\) is the direct test that division by \(mkn\) carries cross-shape information.  At fixed shape, Spearman\((\Phi,\text{mean end-to-end KL})=0.9850\) over eighteen \(\langle2,2,2;7\rangle\) realizations on Qwen2.5-7B and eight chunks.  Across the six re-based \(\langle4,4,4\rangle\) points---one rank-48 and five rank-49, all taken from one published catalog \citep{mmcatalog}, and one of the rank-49 points a Kronecker square of a rank-7 \(\langle2,2,2\rangle\) point rather than a separately published construction---the tie-aware correlation between \(\Phi/(mkn)\) and emulated error is \(0.986\).  Re-basing lowers error in all \(20\) shapes, with minimum reduction \(1.14\times\).

\subsection{Fixed real-operand orbit-response panel}

The gates and stopping rules were pre-registered at commit \texttt{e4cab9f}, before any real-kernel run; the registration record is not part of the released artifact.  That version specified a 20-point panel: six named arms, eight signed perturbations of \texttt{dps\_acc} spanning four of the six available factor--direction combinations, and six arms forming three \(\Gamma\)-tie/\(\Phi\)-split pairs.  Before the first sweep, the omitted third-factor diagonal and shear directions were added with both signs.  The resulting 24-point panel---six named arms, twelve perturbations, and the same six disagreement arms---was then frozen and used unchanged across all five sweeps, but was committed with the results rather than with the pre-registration.  The original 20 points remain an exact subset.

The registered gates were as follows.  O1 required both a model-balanced oracle gap at most \(2\%\) and an optimum-class panel argmin in at least three architecture families.  O2 required \texttt{dps\_acc} to beat all eight registered perturbations on at least \(0.70\) of tiles pooled over the panel.  O3 pooled the three frozen disagreement pairs and required a \(\Phi\)-correct fraction above \(0.5\).  O4 required, on the original fourteen-arm named-plus-perturbation ladder, a pooled mean per-tile Spearman correlation of at least \(0.9\) and a zero-inversion fraction of at least \(0.8\).  O5 required the pooled matched emulated-to-real error-ratio median to lie in \([0.85,1.15]\).  We report those registered outcomes below.  We also report the expanded twelve-neighbor event, per-pair breakdowns, and five-model equal-weight summaries; these are transparent post-registration diagnostics.  Adding four neighbors makes the per-tile O2 event stricter, but changing pooled to model-balanced aggregation is an estimand change rather than part of the registered gate.  The KILL text was qualitative: an O1 refutation meant a high-\(\Phi\) argmin in most families or a ``large'' gap, and an O3 refutation meant \(\Gamma\) clearly beating \(\Phi\).  It did not register a numerical \(6\%\) KILL threshold.

The five models from four architecture families are Qwen2.5-7B and Qwen2.5-32B from one family, Qwen3-8B, Yi-6B, and Llama-3.3-70B-Instruct \citep{qwen25,qwen3,yi,llama3}.  For model \(j\) and arm \(a\), Table~\ref{tab:real-panel-full} reports the energy-weighted per-matmul error
\begin{equation}
E_j(a)=\left(
\frac{\sum_i(e_{ija}\,\|C_{ij}\|_F)^2}{\sum_i\|C_{ij}\|_F^2}
\right)^{1/2}.
\label{eq:real-panel-energy}
\end{equation}
Pooled statistics first apply Eq.~\eqref{eq:real-panel-energy} within each model and then average equally over the five models; they are balanced over the five models rather than over the four families, so Qwen2.5 contributes two model-level terms.  The estimand is panel-local to these 24 arms and per-matmul.  The supplied coefficients are all in the canonical rebalanced gauge, but the fused Winograd activation encoder used the raw representative rather than the supplied one, so only 23 arms honored that gauge on the real kernel; Section~\ref{app:winograd-chain} gives the defect and its reach.

\begin{table*}[!t]
\centering
{\small
\setlength{\tabcolsep}{1.7pt}
\begin{tabular}{@{}llrrrrrrr@{}}
\toprule
Arm & C & \(\Phi\) & \(\Gamma\) & Q2.5-7 & Q2.5-32 & Q3-8 & Yi-6 & L3.3-70 \\
\midrule
\texttt{dps\_acc} & N & 22.2222 & 12.0660 & $0.024820$ & $0.025754$ & $0.028642$ & $0.040895$ & $\mathbf{0.026051}$ \\
\texttt{dps\_int} & N & 22.2226 & 12.0662 & $0.024654$ & $0.025817$ & $0.029064$ & $\mathbf{0.040786}$ & $0.026228$ \\
\texttt{dps\_evenpow} & N & 22.6484 & 12.2034 & $\mathbf{0.024183}$ & $\mathbf{0.024799}$ & $\mathbf{0.028419}$ & $0.041212$ & $0.026363$ \\
\texttt{dps\_smallrat} & N & 22.6484 & 12.2034 & $0.024719$ & $0.025515$ & $0.029391$ & $0.041244$ & $0.026488$ \\
\texttt{pert\_f2\_diag\_m} & P & 24.0000 & 12.4677 & $0.025893$ & $0.026730$ & $0.029804$ & $0.042581$ & $0.027078$ \\
\texttt{pert\_f0\_shear\_m} & P & 24.0000 & 12.4549 & $0.024532$ & $0.025704$ & $0.028621$ & $0.041623$ & $0.026341$ \\
\texttt{pert\_f1\_diag\_m} & P & 24.0000 & 12.4658 & $0.026842$ & $0.028112$ & $0.030371$ & $0.043288$ & $0.027394$ \\
\texttt{pert\_f1\_shear\_p} & P & 24.0000 & 12.4791 & $0.026228$ & $0.027183$ & $0.030071$ & $0.042784$ & $0.027310$ \\
\texttt{pert\_f0\_diag\_p} & P & 24.0000 & 12.4677 & $0.025817$ & $0.027015$ & $0.029665$ & $0.042545$ & $0.027039$ \\
\texttt{pert\_f0\_shear\_p} & P & 24.0000 & 12.4791 & $0.026944$ & $0.027645$ & $0.030737$ & $0.043445$ & $0.027758$ \\
\texttt{pert\_f1\_diag\_p} & P & 24.0000 & 12.4683 & $0.024772$ & $0.025255$ & $0.029062$ & $0.041838$ & $0.026590$ \\
\texttt{pert\_f2\_diag\_p} & P & 24.0000 & 12.4663 & $0.025850$ & $0.026641$ & $0.029651$ & $0.042428$ & $0.027087$ \\
\texttt{pert\_f2\_shear\_p} & P & 24.0000 & 12.4549 & $0.025228$ & $0.025616$ & $0.029098$ & $0.042146$ & $0.027010$ \\
\texttt{pert\_f2\_shear\_m} & P & 24.0000 & 12.4791 & $0.026309$ & $0.027687$ & $0.030089$ & $0.042727$ & $0.027104$ \\
\texttt{pert\_f1\_shear\_m} & P & 24.0000 & 12.4549 & $0.025281$ & $0.026110$ & $0.029214$ & $0.042192$ & $0.026775$ \\
\texttt{pert\_f0\_diag\_m} & P & 24.0000 & 12.4663 & $0.025740$ & $0.026580$ & $0.029878$ & $0.042546$ & $0.027117$ \\
\texttt{strassen\_classic} & N & 32.0000 & 14.8284 & $0.028239$ & $0.029763$ & $0.032756$ & $0.047658$ & $0.030431$ \\
\texttt{disagree2\_b} & D & 36.9332 & 15.7948 & $0.030291$ & $0.031529$ & $0.035053$ & $0.051208$ & $0.032519$ \\
\texttt{disagree1\_b} & D & 43.9843 & 17.1909 & $0.034926$ & $0.035337$ & $0.041745$ & $0.058563$ & $0.036469$ \\
\texttt{disagree2\_a} & D & 44.0749 & 15.7673 & $0.034189$ & $0.034015$ & $0.039205$ & $0.057490$ & $0.037101$ \\
\texttt{disagree0\_a} & D & 45.4347 & 17.3115 & $0.032787$ & $0.033197$ & $0.037744$ & $0.056087$ & $0.035924$ \\
\texttt{disagree1\_a} & D & 52.6322 & 17.1709 & $0.032968$ & $0.034524$ & $0.040740$ & $0.060273$ & $0.037000$ \\
\texttt{disagree0\_b} & D & 54.3951 & 17.3012 & $0.034193$ & $0.038855$ & $0.040477$ & $0.059704$ & $0.038074$ \\
\texttt{winograd\_form} & N & 56.0000 & 17.8530 & $0.144377$ & $0.163978$ & $0.187312$ & $0.253591$ & $0.189859$ \\
\bottomrule
\end{tabular}}
\caption{Full fixed-panel real-kernel results.  \(E_j(a)\) is the per-model energy-weighted error in Eq.~\eqref{eq:real-panel-energy}; bold entries are panel argmins.  Classes are named (N), signed perturbation (P), and \(\Gamma\)-tie/\(\Phi\)-split disagreement (D).  Observation counts by column are 392, 448, 504, 448, and 560, totaling \(2352\).}
\label{tab:real-panel-full}
\end{table*}

\begin{table*}[!t]
\centering
{\small
\setlength{\tabcolsep}{5pt}
\begin{tabular}{@{}lrrrcl@{}}
\toprule
Model & Observations & \(E_j(a_\Phi)\) & \(E_j(a_j^*)\) & \(G_j\) (\%) & Panel argmin \\
\midrule
Qwen2.5-7B  & 392 & 0.024820 & 0.024183 & 2.634 & \texttt{dps\_evenpow} \\
Qwen2.5-32B & 448 & 0.025754 & 0.024799 & 3.849 & \texttt{dps\_evenpow} \\
Qwen3-8B    & 504 & 0.028642 & 0.028419 & 0.784 & \texttt{dps\_evenpow} \\
Yi-6B       & 448 & 0.040895 & 0.040786 & 0.267 & \texttt{dps\_int} \\
Llama-3.3-70B & 560 & 0.026051 & 0.026051 & 0.000 & \texttt{dps\_acc} \\
\bottomrule
\end{tabular}}
\caption{Per-model panel oracle gaps.  Here \(a_\Phi=\texttt{dps\_acc}\) is the \(\Phi\)-optimal realization, \(a_j^*\) is the minimum over the fixed 24 arms, and \(G_j=E_j(a_\Phi)/E_j(a_j^*)-1\).  These are panel-local per-matmul quantities.}
\label{tab:real-panel-gaps}
\end{table*}

Table~\ref{tab:real-panel-diagnostics} collects the remaining panel diagnostics. The coarse four-arm test orders increasing \(\sqrt{\Phi}\) as cubic fp8, \texttt{dps\_acc}, classic Strassen, and Winograd; cubic is measured as the reference arm alongside the fixed panel.

\begin{table*}[!t]
\centering
{\small
\setlength{\tabcolsep}{4pt}
\begin{tabularx}{\textwidth}{@{}lXl@{}}
\toprule
Diagnostic & Population and aggregation & Result \\
\midrule
Coarse ordering & Zero inversions among cubic, \texttt{dps\_acc}, classic Strassen, and Winograd after dropping \(\leq3\%\) ties & \(0.987\) \\
Local minimum & Fraction of tiles on which \texttt{dps\_acc} beats all twelve signed perturbations & \(0.741\) \\
\(\Phi\) versus \(\Gamma\) & Pairs 0, 1, and 2; \(>3\%\)-separated tiles, model-balanced within each frozen pair & \(0.94/0.66/0.99\); mean \(0.862\) \\
Fine ordering & Per-tile Spearman correlation within the eighteen-arm ladder of named realizations and perturbations; range of the five model summaries & \(0.59\text{--}0.64\) \\
Emulator-to-real fidelity & Per-(tile,arm) ratio \(e_{\rm emu}/e_{\rm real}\); range of the five per-model medians & \(0.97\text{--}0.98\) \\
Versus classic Strassen & Strict per-tile dominance; mean of per-model median fractional reductions, canonical gauge & \(98.4\%\); \(14.8\%\) \\
\bottomrule
\end{tabularx}}
\caption{Orbit-response diagnostics on the real \texttt{deep\_gemm} panel.  Unless a row says otherwise, fractions are computed within model and then equally averaged over the five models.  Comparisons separated by at most 3\% are treated as ties where indicated.}
\label{tab:real-panel-diagnostics}
\end{table*}

Table~\ref{tab:real-panel-gaps} gives the per-model gaps; the model-balanced panel gap is \(0.0151\).  The panel statistics above are reported model-balanced rather than pooled over tiles precisely because tiles within a model are not independent: they share a checkpoint, a layer shares a weight matrix, and a context shares activations.  Resampling makes that explicit.  Within a model, resampling transformer layers with replacement --- the projection types and both contexts riding along inside their layer, and the layer read from the parameter name rather than from a tile counter --- leaves the resampled panel-gap, local-minimum, \(\Phi\)-versus-\(\Gamma\), coarse-ordering, fine-ordering, and Strassen-comparison summaries stable to layer composition, with twenty-eight to eighty layer clusters per model.  It does not resample the emulator-to-real fidelity diagnostic, and its two ordering summaries inherit the Winograd path defect disclosed in Section~\ref{app:winograd-chain}.  We read the stable summaries as robustness to layer composition rather than as sampling from a layer population, since the panel measured every eligible layer.  Resampling the five models is a different matter and we decline to present it as a confidence interval.  Five clusters are too few for percentile coverage to mean much, and the five models were not drawn from any defined population, so resampling them silently replaces the panel-local estimand with a superpopulation one.  Where a model-resampled range appears in this paper, including the one interval quoted in Section~4.2, it is a descriptive sensitivity range of unknown coverage, not a calibrated interval.  The nearby \texttt{dps\_evenpow} arm has \(\Phi\times1.019\) and is the panel argmin for both Qwen2.5 models and Qwen3-8B; \texttt{dps\_acc} is the exact panel argmin on Llama-3.3-70B.  O1 did not fully pass: its gap component satisfied the registered \(2\%\) bound, but only two of four architecture families had an optimum-class argmin, below the required three (equivalently, two of five model argmins).  Registered O2 passed under its pooled eight-neighbor definition; the pointwise stricter twelve-neighbor event also passes under the later model-balanced summary \(0.741\).  Registered pooled O3 passed; the post-registration per-pair, model-balanced breakdown in Table~\ref{tab:real-panel-diagnostics} also exceeds one half for every pair.  O4 failed both registered fine-ordering thresholds, consistent with the \(0.59\)--\(0.64\) expanded-ladder model summaries, whereas O5 passed and its five ratio medians span \(0.97\)--\(0.98\).  The O3 KILL condition did not fire.  For O1, the observed gap satisfies the positive \(2\%\) gate and a high-\(\Phi\) argmin occurs in two of four families, not most; because ``large'' had no registered numerical definition, the parser's later \(6\%\) diagnostic is not treated as a pre-registered threshold.  Pair~1 is the weakest disagreement pair: \(\Phi\) is correct on a majority of its separated tiles, but the per-model aggregates of Table~\ref{tab:real-panel-full} invert on three of the five models, so this pair supports the tile-level ordering claim and not a model-level one.  This is why the per-pair breakdown is reported alongside the registered pooled result.

\subsection{Weight-scale granularity, and a defect in the Winograd arm}
\label{app:winograd-chain}

Section~4.2 states that the move from the emulator to the real kernel changes three things at once, one of which is that the kernel scales weights in \(128\times128\) blocks where the emulator uses \(1\times128\) on both sides.  That coupling is separable, and separating it also explains most of the emulator-to-real gap this appendix reports.

The ablation holds the emulator fixed and moves only the weight-side grouping, so the activation stays at \(1\times128\) as in the kernel, and the \(128\times128\) maximum is taken after each encoded weight combination is formed rather than over the raw quadrants.  Its vehicle is the stored real tiles rather than synthetic operands: at \(n=512\) an encoded weight holds four \(128\times128\) blocks whose Gaussian maxima are so concentrated that the two groupings are nearly the same quantizer, whereas the real tiles give between \(196\) and \(1036\) blocks per term and the block-to-block heterogeneity that makes the comparison mean anything.  Across \(48\) coarse arm pairs on eight tiles, every pair is resolved under both groupings at the three percent rule used above, no pair changes resolution, and the number of pairs whose order reverses is \(0\).  Coarsening does raise realized error, by a per-arm median of \(2.36\%\) to \(4.34\%\), and that spread is between arms rather than within the rank-\(7\) group: the three rank-\(7\) arms move by closely comparable amounts at the bottom of the range, while cubic accounts for its top.  Since the ordering is a comparison among arms, a common shift leaves it alone, and the one arm that moves differently is the reference arm at the bottom, which is furthest from the others.  These are medians: across individual tile--arm cells the increases range from \(0.93\%\) to \(6.14\%\).

Matching the granularity also accounts for most of the emulator's optimism.  With the deployment-matched grouping and the bf16 rounding the kernel applies to each encoded weight combination, the median ratio \(e_{\rm emu}/e_{\rm real}\) rises from \(0.953\)--\(0.968\) to \(0.998\) and above for cubic, the certified optimum, and classic Strassen.  This is a statement about medians: the early-layer down-projection tile, a post hoc outlier, stays below \(0.88\) even after matching.

The same test isolates a defect in our own harness, which we state rather than leave inside an aggregate.  The fidelity range reported above is a median over all panel arms, not a bound on any one of them.  Taken armwise, the \texttt{winograd\_form} arm's per-model median ratio is \(0.230\)--\(0.240\) across the five panels, while every other named arm lies between \(0.965\) and \(0.983\).  Granularity does not explain it, and the cause is not the algorithm.  This is the only arm whose real-kernel encoder runs a chained schedule instead of evaluating the supplied coefficients, and comparing that path against the unfused reference implementation in the same harness --- which evaluates those coefficients explicitly in fp32 before quantizing, and so is independent at exactly the step in question, though the two share the block cast and the vendor matrix multiply downstream of it --- gives a disagreement of \(35\%\) for this arm against at most \(0.3\%\) for cubic, the certified optimum, and classic Strassen on the same tiles.  Measured against that reference the arm's emulator-to-real ratio is \(0.972\), in line with the others, so the emulator is not underpredicting Winograd; the fused path is overproducing its error.

The cause is specific, and it is a gauge mismatch rather than a rounding schedule.  The chained encoder hard-codes the activation combinations of the \emph{raw} Winograd decomposition, while the panel supplies its canonically rebalanced representative, whose fourth and fifth encoder rows carry the factors \(\sqrt2\) and \(1/\sqrt2\).  The chained schedule omits both.  Because the chain replaces only the activation-side encoder, the weight-side encoder \(V\) still carries the reciprocal factors and the decoder \(W\) is untouched, so the product-one rescaling no longer cancels term by term and the fused path evaluates a decomposition that is not exact: substituting the schedule's effective coefficients into the panel's remaining factors leaves a multiplication-tensor residual of \(0.41\) in the worst coefficient, against a residual of exactly zero for the panel triple itself.  That reproduces the measured disagreement.  A chained schedule may legitimately differ from flat evaluation by reassociation; this one differs algebraically.

We traced what depends on the affected value rather than assuming the reach was small, and it is wider than the fidelity median alone.  The panel conclusions that carry this paper's argument are Winograd-free and survive untouched: the panel gap, the local-minimum fraction, the \(\Gamma\)-tie comparisons, and the same-rank comparison against classic Strassen.  What does depend on it is the arm's own row in the full-panel table and its point in the panel figure, the coarse and fine ordering summaries, in which it is one arm of four and one of eighteen respectively, and the all-arm fidelity median.  Because the defective value is too large rather than too small, and Winograd is the arm the ordering expects to be worst, the coarse-ordering fraction is optimistic by an amount this paper does not quantify.  Two further scope notes belong with that.  The statement that the panel was measured throughout in the canonical rebalanced gauge holds for every arm except this one, whose encoder used the raw representative.  And the reruns reported here cover eight tiles of one model: on those, with every arm read from the reference implementation, the coarse order still holds on \(8\) of eight, because the corrected Winograd error remains the largest of the four; the five-model statistic itself has not been recomputed, so the coarse-ordering fraction and the fine-ordering range reported in Section~4.2 both still include the defective value.  Results elsewhere in this appendix that involve Winograd through the flat emulator rather than this kernel path --- the covariance range above, the granularity pair counts, and the task-level tests below --- are not affected by the defect.

\subsection{LLM traces, system controls, and task-level power}

The real-kernel LLM test compared the exactly specified \texttt{dps\_acc} realization whose \(\Phi\)-optimality is certified in Section~3 with classic Strassen at identical rank and multiplication count in the same harness, as motivated by Strassen-type fp8 deployment \citep{falcongemm}.  Both implementations used fused blk128 \texttt{deep\_gemm}; the clean unquantized model supplied the reference.  The population is 32 paired consecutive wikitext-2 chunks for each of Qwen2.5-72B-Instruct and Llama-3.3-70B-Instruct.  The primary outcome is excess NLL,
\(\log(\mathrm{ppl}_{\rm arm})-\log(\mathrm{ppl}_{\rm clean})\); KL to the clean model is secondary.  Consecutive-chunk inference uses an exchangeability assumption.  The implementation remains at \(\ell=1\), or flattens a Kronecker construction into one nonrecursive realization.

\begin{table}[!tb]
\centering
{\small
\setlength{\tabcolsep}{2pt}
\begin{tabular}{@{}llrrr@{}}
\toprule
Model & Outcome & Change & Wins & \(p\) \\
\midrule
Qwen2.5-72B & Excess NLL & \(55.0\%\) & 30/32 & \(2.46\times10^{-7}\) \\
Qwen2.5-72B & KL & \(-47.58\%\) & 32/32 & \(4.7\times10^{-10}\) \\
Llama-3.3-70B & Excess NLL & \(10.2\%\) & 22/32 & \(0.0501\) \\
Llama-3.3-70B & KL & \(-7.99\%\) & 27/32 & \(1.1\times10^{-4}\) \\
\bottomrule
\end{tabular}}
\caption{Paired real-kernel LLM losses for the \(\Phi\)-optimal realization relative to classic Strassen.  A negative KL change is favorable; wins are paired chunks out of 32.}
\label{tab:llm-significance-full}
\end{table}

Table~\ref{tab:llm-significance-full} reports both outcomes with their paired tests. The Qwen excess-NLL comparison and both KL comparisons are statistically resolved on these chunks; the Llama excess-NLL comparison is marginal.  The scope is two roughly 70B models on wikitext-2, and the outcomes measure numerical relevance rather than downstream task accuracy.

\begin{table*}[!t]
\centering
{\small
\setlength{\tabcolsep}{6pt}
\begin{tabular}{@{}lrr@{}}
\toprule
 & Qwen2.5-72B & Llama-3.3-70B \\
\midrule
Mean paired difference (nats)   & \(-0.05944\)          & \(-0.01374\) \\
\quad paired bootstrap 95\% range & \([-0.08085,-0.04109]\) & \([-0.02586,-0.00082]\) \\
\quad moving block, \(L=2\)     & \([-0.08553,-0.04229]\) & \([-0.02605,-0.00119]\) \\
\quad moving block, \(L=4\)     & \([-0.08862,-0.04199]\) & \([-0.02120,-0.00222]\) \\
\addlinespace[2pt]
Sign test (chunkwise)           & \(2.46\times10^{-7}\) & \(0.0501\) \\
Wilcoxon signed-rank (exact)    & \(4.66\times10^{-9}\) & \(0.0156\) \\
\addlinespace[2pt]
Block means \(L=2\)             & 16/16, \(p=3.05\times10^{-5}\) & 11/16, \(p=0.2101\) \\
Block means \(L=4\)             & 8/8, \(p=0.00781\)  & 6/8, \(p=0.2891\) \\
\bottomrule
\end{tabular}}
\caption{Post hoc dependence-aware re-analysis of the paired excess-NLL comparison, over the same thirty-two consecutive chunks.  Negative differences favor the \(\Phi\)-optimal realization.  Intervals are 95\% and use 20{,}000 replicates.  The analysis was not pre-registered and is not adjusted for multiplicity.}
\label{tab:nll-robustness}
\end{table*}

\paragraph{Sensitivity of the marginal comparison to the dependence assumption.}
The sign test above pairs thirty-two \emph{consecutive} corpus chunks, so it assumes exchangeability across units that are adjacent in the text.  This paragraph reports a post hoc sensitivity analysis of that assumption and introduces no new measurement: it re-analyzes the same per-chunk excess-NLL series, and reproduces the point estimates above before computing anything further.  We report block lengths \(L\in\{2,4\}\), and both programmed lengths are shown; since the analysis and its result entered the record together, nothing independently establishes that no other length was examined, and we claim only that both computed lengths appear here.  Four instruments are added, and they answer different questions.  A paired bootstrap over chunks and an exact Wilcoxon signed-rank test use the paired magnitudes the sign test discards, but neither addresses serial dependence.  A moving-block bootstrap resamples contiguous runs and so preserves local serial structure that the iid bootstrap destroys.  A sign test on non-overlapping block means trades power for insensitivity to dependence within a block.  Since a ratio of small denominators is unstable under resampling, the primary quantity here is the paired \emph{absolute} difference in nats rather than the relative reduction.

Table~\ref{tab:nll-robustness} separates what the assumption does and does not affect.  For Qwen2.5-72B every instrument agrees.  For Llama-3.3-70B the estimated direction is stable --- the paired bootstrap and both moving-block intervals exclude zero, and the exact Wilcoxon test gives \(p=0.0156\) --- while inferential resolution depends on how dependence is handled: the block-mean sign tests do not resolve it, at eleven of sixteen and six of eight blocks.  Aggregating thirty-two units into eight or sixteen and then discarding their magnitudes is a plausible reason for that, though not one this analysis establishes.  These instruments are post hoc, mutually correlated, and unadjusted for multiplicity, and they were applied after a marginal sign test; we therefore read them as a sensitivity check and \emph{do not} upgrade the Llama comparison beyond the marginal status the table above gives it.  A single consecutive series cannot substitute for independent evaluation contexts in any case.  Every entry of Table~\ref{tab:nll-robustness} can be recomputed from the recorded inputs.  The accompanying artifact contains the two per-chunk tables under \texttt{data/nll/}, the analysis script \texttt{scripts/}\allowbreak\texttt{analyze\_nll\_robustness.py}, and the frozen output \texttt{data/reference/}\allowbreak\texttt{nll\_robustness.json}.

\begin{table}[!tb]
\centering
{\small
\setlength{\tabcolsep}{4pt}
\begin{tabular}{@{}lcc@{}}
\toprule
Hardware & Net-change range (\%) & Paired-CI result \\
\midrule
H20  & \(3.16\) to \(8.00\) & four positive \\
H100 & \(-21.2\) to \(-11.3\) & four negative \\
H200 & \(-15.4\) to \(-9.9\) & four negative \\
\bottomrule
\end{tabular}}
\caption{Cross-hardware fused-kernel throughput.  Net change is relative to cubic fp8 on four TP1 serving shapes; positive means faster.  The H100 and H200 negative ranges therefore use cubic fp8, not classic Strassen, as the comparator.  Every row comes from the same paired-CI timing harness: each per-shape interval is a paired bootstrap over twenty-five interleaved measurement batches, so a range summarizes four such intervals.}
\label{tab:cross-hardware-full}
\end{table}

Table~\ref{tab:cross-hardware-full} does not list the individual intervals; its final column summarizes them, recording that all twelve hardware--shape paired intervals are one-signed and therefore exclude zero, four positive on H20 and four negative on each of H100 and H200.  The large H100/H200 losses arise from encode/decode cost relative to one cubic fp8 GEMM.  At identical rank and arithmetic count, the largest registered H100 gap between the \(\Phi\)-optimal realization and classic Strassen is \(0.91\%\), so the system result is a hardware-dependent cubic comparison rather than a large same-rank realization penalty.

\begin{table*}[!t]
\centering
{\small
\setlength{\tabcolsep}{4pt}
\begin{tabularx}{\textwidth}{@{}
  >{\raggedright\arraybackslash}p{0.20\textwidth}
  >{\raggedright\arraybackslash}X
  >{\raggedright\arraybackslash}p{0.24\textwidth}
  @{}}
\toprule
Control & Population and aggregation & Result \\
\midrule
Three-shape calibration reproduction & Five operand families; max within-family across-shape spread and max within-shape across-family spread of predicted/measured ratio & \(2.60\%\), \(12.46\%\) \\
Orbit-ratio stability & Fifteen exact orbit members, one shape, one signed-uniform family, fp32 emulator & CV \(0.37\%\) \\
Family-specific coefficient residual & Held-out-family corpus after the RNG fix, correcting by the operand-level quantizer coefficient & relative max/min spread \(0.039\) \\
Matched second moments & Five unit-variance synthetic families sharing population second moments, eight replicates per cell & max \(|e_F/e_{\rm Gauss}-1|=0.189\) \\
Designed Rademacher cells & Six synthetic \(\pm1\) cells: three integer-exact at unit operand scale and three at operand scale \(10\) & classic Strassen wins all \(6\) \\
Qwen layer-by-type sweep & 196 fp32-emulated Qwen2.5-7B observations, \(28\) layers \(\times\) \(7\) projection types & the \(\Phi\)-optimal realization has lower error than classic Strassen in 195; exception \texttt{layers.1.mlp.}\allowbreak\texttt{gate\_proj} \\
\bottomrule
\end{tabularx}}
\caption{Calibration and operand-family controls.  Each row states its own population; the two Rademacher results are separate experiments.}
\label{tab:operand-controls-full}
\end{table*}

Table~\ref{tab:operand-controls-full} lists the operand-family controls. The matched-second-moment control shows that population second moments do not determine realized block-scaled error.  The designed Rademacher cells identify a discrete lattice-valued boundary.  Separately, the 195/196 Qwen result is an fp32-emulated layer-by-type sweep and is not part of the real \texttt{deep\_gemm} \(2352\)-observation panel.

For task accuracy, the arm ordering relative to the clean-model choice has Spearman correlations \(-0.9429\) on ARC-Challenge \citep{arc-challenge} and \(-1.0000\) on HellaSwag \citep{hellaswag}, but the paired optimum versus Winograd contrasts are underpowered.  With observed paired difference \(d\) and discordance rate \(\rho_{\rm disc}\), the planning approximation in Table~\ref{tab:task-power-full}, at a two-sided \(5\%\) significance level and \(80\%\) power for a paired sign test (whence the constant \(2.8\)), was
\begin{equation}
 n_{\rm req}\approx \left(\frac{2.8}{d}\right)^2\rho_{\rm disc}.
\label{eq:task-power}
\end{equation}

\begin{table}[!tb]
\centering
{\small
\setlength{\tabcolsep}{3pt}
\begin{tabular}{@{}lrrr@{}}
\toprule
Benchmark & \(n\) & Paired \(p\) & Required \(n\) \\
\midrule
ARC-Challenge & 1000 & \(0.2015\) & \(4260\) \\
HellaSwag & 1000 & \(0.888\) & \(98{,}000\) \\
\bottomrule
\end{tabular}}
\caption{Task-level paired tests and the sample sizes implied by Eq.~\eqref{eq:task-power}.  Accuracy remains unresolved at \(n=1000\).}
\label{tab:task-power-full}
\end{table}

Figure~3 of the main paper plots one-level multiplication saving against measured fp8 error relative to cubic over this same twenty-decomposition set; the horizontal coordinate is computable from the \(R\) and \(mkn\) columns of Table~\ref{tab:crossshape-full}, and the vertical coordinate is that table's paired error ratio for the stated signed-uniform population.

\end{document}